\documentclass{article} 
\usepackage[final]{colm2026_conference}

\usepackage{microtype}
\usepackage{hyperref}
\usepackage{url}
\usepackage{booktabs}

\usepackage{lineno}

\definecolor{darkblue}{rgb}{0, 0, 0.5}
\definecolor{Gray}{gray}{0.93}
\definecolor{highlight}{HTML}{E0E0FF}
\hypersetup{colorlinks=true, citecolor=darkblue, linkcolor=darkblue, urlcolor=darkblue}

\usepackage{graphicx}
\usepackage{wrapfig}

\usepackage{algorithm}
\usepackage{algorithmic}
\usepackage{amssymb}

\usepackage{tabularx}
\usepackage{array}
\usepackage{fontawesome5}
\usepackage{mathtools}
\usepackage{caption}

\usepackage{amsfonts}       
\usepackage{nicefrac}       
\usepackage{microtype}      
\usepackage{xcolor}         

\usepackage{subcaption}
\usepackage{amsmath}
\usepackage{multirow}
\usepackage[table]{xcolor}
\usepackage{adjustbox}
\usepackage{diagbox}
\usepackage{makecell}

\newcommand{\methodname}{DR-LoRA}

\newcommand{\olmoe}{OLMoE}
\newcommand{\llamamoe}{LLaMA-MoE}

\newcommand{\gsm}{GSM8K}
\newcommand{\humaneval}{HumanEval}
\newcommand{\ifeval}{IFEval}

\newcommand{\medmcqa}{MedMCQA}

\newcommand{\wmt}{WMT23}
\newcommand{\ledgar}{LEDGAR}

\title{\methodname: Dynamic Rank LoRA for Fine-Tuning\\ Mixture-of-Experts Models}

\author{%
\parbox{0.96\textwidth}{%
\raggedright
{\small
\textbf{Guanzhi Deng$^{1,2*}$}\enspace
\textbf{Bo Li$^{3*}$}\enspace
\textbf{Ronghao Chen$^{4*}$}\enspace
\textbf{Xiujin Liu$^{5}$}\enspace
\textbf{Zhuo Han$^{1,2}$}\\[-0.15em]
\textbf{Huacan Wang$^{4}$}\enspace
\textbf{Lijie Wen$^{3\dagger}$}\enspace
\textbf{Linqi Song$^{1,2\dagger}$}
}\\[0.25em]
{\normalfont\footnotesize
$^{1}$ City University of Hong Kong\\[-0.12em]
$^{2}$ City University of Hong Kong Shenzhen Research Institute\\[-0.12em]
$^{3}$ Tsinghua University\enspace
$^{4}$ QuantaAlpha\enspace
$^{5}$ University of Michigan
}\\[0.15em]
}%
}

\begin{document}

\ifcolmsubmission
\linenumbers
\fi

\maketitle

\begin{abstract}
Mixture-of-Experts (MoE) has become a prominent paradigm for scaling Large Language Models (LLMs). Parameter-efficient fine-tuning methods, such as LoRA, are widely adopted to adapt pretrained MoE LLMs to downstream tasks. 
However, existing approaches typically assign identical LoRA ranks to all expert modules, ignoring the heterogeneous specialization of pretrained experts. 
This uniform allocation leads to a resource mismatch: task-relevant experts are under-provisioned, while less relevant ones receive redundant parameters. 
To address this, we propose \textbf{\methodname}, a \textbf{D}ynamic \textbf{R}ank \textbf{LoRA} framework for fine-tuning pretrained MoE models. 
Specifically, \methodname~initializes all expert LoRA modules with a small active rank and uses an expert saliency score, which combines routing frequency and gradient-based rank importance, to identify which experts would benefit most from additional capacity. 
It then periodically expands the active ranks of the task-critical expert LoRA, progressively constructing a heterogeneous rank distribution tailored to the target task. 
Experiments on three MoE models across six tasks show that \methodname~consistently outperforms LoRA and other strong baselines, demonstrating that task-adaptive heterogeneous rank allocation is an effective strategy to improve active capacity utilization in MoE fine-tuning.

\vspace{0.1cm}
\centering{\textbf{
\faGithub~ Code: \url{https://github.com/gz-d/dr-lora}
}}
\end{abstract}

\section{Introduction}
\let\thefootnote\relax
\footnotetext{%
$^{*}$Equal Contribution.\quad
$^{\dagger}$Corresponding Authors.%
}
Mixture-of-Experts (MoE) has become a prominent paradigm for scaling Large Language Models (LLMs) \citep{yang2025qwen3,jiang2024mixtral, liu2024deepseek}.
By activating only a subset of experts for each input token, MoE substantially increases model capacity without proportionally increasing per-token computation, demonstrating impressive capabilities across a wide range of tasks~\citep{shazeer2017,li-etal-2026-macs,team2025kimi,li2026mix}.
With the widespread adoption of MoE LLMs, efficiently adapting them to specific downstream tasks has become a significant challenge.

Parameter-Efficient Fine-Tuning (PEFT), particularly LoRA~\citep{hu2022lora}, is a primary approach to address this challenge.
However, when applied to MoE models, existing methods typically assign a LoRA rank to all 
experts~\citep{li2024mixlora,liu2024moe,dou2024loramoe,gao2025mola,liu2024perft}, implicitly assuming that all experts have similar adaptation demands. 
This assumption overlooks the heterogeneous activation patterns and functional specialization that experts develop during pretraining~\citep{wang2025hmoe,fedus2022switch,jiang2024mixtral,dai2024deepseekmoe}.
Uniform rank allocation thus leads to \textbf{Expert-Level Capacity Mismatch}: high-frequency experts lack sufficient parameters to fully adapt, while low-frequency experts occupy capacity that contributes little to task performance.

Previous work provides partial solutions but does not fully address the capacity mismatch problem. 
For example, PERFT-R~\citep{liu2024perft} introduces routable adaptation modules to improve MoE fine-tuning but uses uniform rank configurations for all experts, without differentiating based on actual usage. 
AdaLoRA~\citep{zhang2023adaptive} achieves dynamic rank allocation through importance-driven pruning but is designed for dense models. It evaluates each LoRA module independently based on gradient signals, without leveraging the routing structure that distinguishes frequently used experts from those rarely activated. 
Additionally, its top-down pruning approach is structurally mismatched with MoE's sparse routing. 
In such sparse settings, low-frequency experts receive only sparse and noisy gradient signals during training, making their importance estimates unreliable and the resulting pruning decisions vulnerable to noise.

\begin{figure*}
    \centering
    \includegraphics[width=0.95\linewidth]{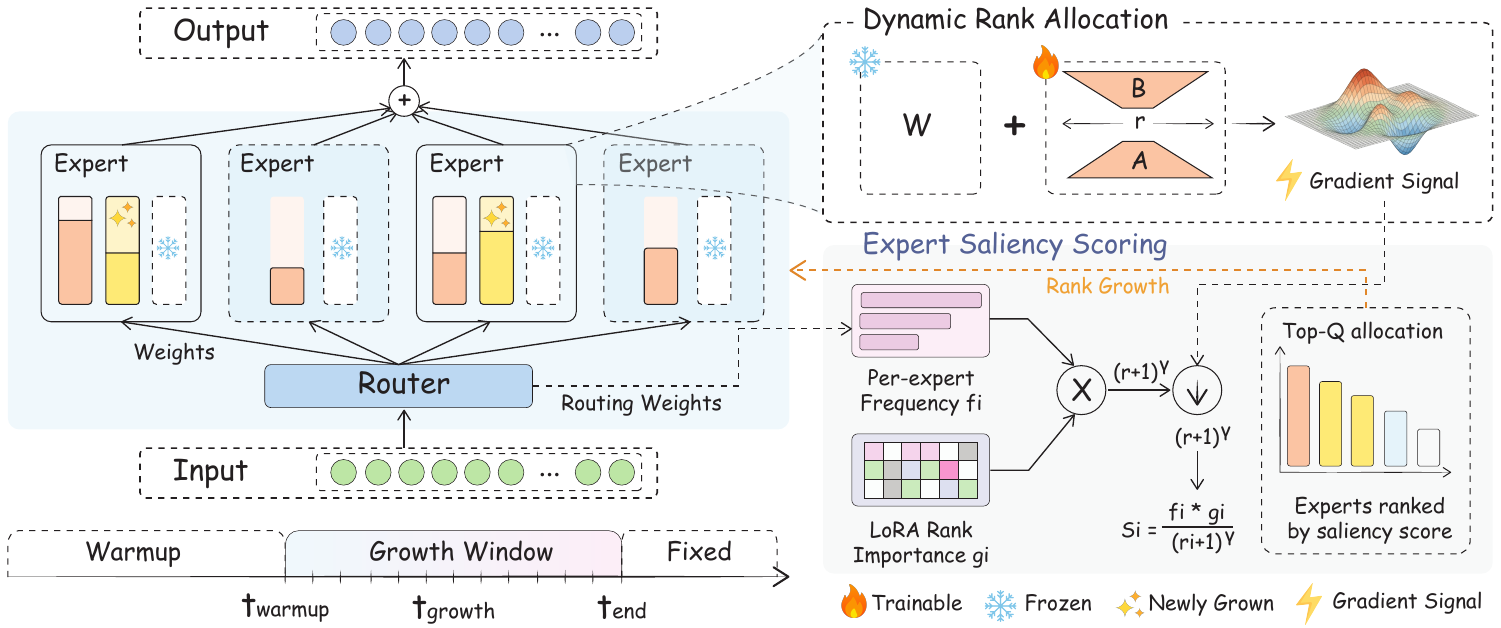}
\caption{Overview of \methodname. Pre-trained expert weights are frozen; each expert's LoRA module starts at a small initial rank and grows dynamically during training. Expert Saliency Scoring integrates routing frequency and rank importance to quantify each expert's demand for capacity expansion. Dynamic Rank Allocation periodically expands active ranks for high-saliency experts, progressively forming a task-adaptive heterogeneous rank structure.}
\label{fig:overview}
\vspace{-5mm}
\end{figure*}

These observations suggest two key requirements for effective MoE fine-tuning:
\textbf{(I)} Rank allocation should reflect the heterogeneous usage patterns of experts on the target task.
\textbf{(II)} The capacity adjustment process must accommodate the sparse and progressive nature of MoE training signals.
To address this, we propose \textbf{\methodname}, a growth-based dynamic rank allocation framework for MoE fine-tuning. 
As shown in Figure~\ref{fig:overview}, \methodname~achieves task-adaptive heterogeneity through two core components.
First, Expert Saliency Scoring combines routing frequency and gradient-based rank importance to score experts for rank growth. 
This module prioritizes experts that are both frequently routed and still actively learning. 
Second, Dynamic Rank Allocation initializes all experts with a small active rank and periodically expands the ranks of high-score experts, avoiding the unreliable early pruning decisions that can arise in top-down methods under sparse routing.
We evaluated \methodname~on three MoE models across six tasks. 
The results show that it consistently outperforms LoRA and other strong PEFT baselines.

The contributions of this work are summarized as follows:
\textbf{(I)} We identify an expert-level capacity mismatch in MoE fine-tuning, where LoRA rank allocation overlooks the heterogeneous expert usage caused by sparse routing, under-provisioning high-frequency experts while wasting capacity on low-frequency experts.
\textbf{(II)} We propose \methodname, a growth-based dynamic rank allocation framework for effective MoE fine-tuning by combining routing frequency and gradient-based rank importance, and a growth-based allocation mechanism for sparse routing.
\textbf{(III)} We validate the effectiveness of our method on three MoE models. Further analysis confirms that \methodname~effectively concentrates the trainable parameter budget on task-critical experts, resulting in a heterogeneous rank distribution.

\section{Related Work}

Parameter-Efficient Fine-Tuning (PEFT) aims to efficiently adapt LLMs by introducing a small number of trainable parameters. 
LoRA \citep{hu2022lora} and its subsequent improvements \citep{liu2024dora,hayou2024lora+} are primarily designed for dense models and do not account for the heterogeneous expert activation patterns introduced by MoE routing.
With the development of MoE models, several works begin to explore the application of LoRA to MoE models \citep{tang2026exploring,pirchert2026flexmore}. 
Typical methods directly inject LoRA into each expert and use a uniform rank allocation strategy. 
However, these methods implicitly assume that all experts require similar fine-tuning capacity, overlooking the task-dependent imbalance in expert activation induced by routing~\citep{dai2024deepseekmoe,wang-etal-2024-expert}. \citep{dai2024deepseekmoe,wang-etal-2024-expert}.
FlexMoRE~\citep{pirchert2026flexmore} investigates rank-heterogeneous experts in federatedly trained MoE and finds that reasoning-heavy tasks benefit from higher expert ranks, providing independent evidence for capacity mismatch. 
Other works introduce MoE structures as plugins in dense models, such as MoLA \citep{gao2025mola}, MiLoRA \citep{li2024mixlora}, CoMoE \citep{feng-etal-2025-comoe}, and LoRAMoE \citep{dou2024loramoe}. 
These methods construct routing mechanisms from scratch to expand the model capacity, which differs fundamentally from fine-tuning of pretrained MoE experts, as studied in this paper.
In recent research, PERFT \citep{liu2024perft} introduces a unified PEFT framework for MoEs, enhancing model expressiveness by building routable adaptation modules (e.g., PERFT-R). 
Although these methods make use of MoE routing, their primary focus is on the design and integration of adaptation modules, rather than on fine-grained capacity reallocation across the original experts.
AdaLoRA \citep{zhang2023adaptive} achieves adaptive rank allocation through singular value pruning. 
This method is effective in dense models, and budgets are allocated using module-level gradient signals. 
However, its importance evaluation assumes that all modules receive sufficient gradient signals, and its top-down pruning process requires all modules to begin training at a high rank. 
Under the sparse routing of MoEs, low-frequency experts accumulate only sparse and noisy gradients, making their importance estimates unreliable.

Unlike previous works, \methodname~uses both routing frequency and gradient-based rank importance to guide rank allocation. 
By adopting a growth-based mechanism that matches the gradual emergence of expert-specific learning signals under sparse routing, it enables fine-grained dynamic capacity allocation at the expert level.

\section{Methodology}
\label{sec:method}

In this section, we present the \methodname~framework, which dynamically allocates expert LoRA ranks during fine-tuning. 
The architecture overview is shown in Figure \ref{fig:overview}. 
We first introduce the problem setting and the motivation for the rank allocation rule (\S\ref{sec:prelim}), then describe Expert Saliency Scoring (\S\ref{sec:saliency}) and Dynamic Rank Allocation (\S\ref{sec:allocation}).


\subsection{Preliminaries and Problem Formulation}
\label{sec:prelim}

We consider a pretrained MoE LLM with $L$ transformer layers, each containing $N$ experts and a router $G_\ell$. 
For an input token $x$, the router computes routing weights and activates the top-$k$ experts. We denote the post-top-$k$ routing weight as $z_{\ell,i}(x)$ (zero when expert $i$ is not selected), so that $\mathrm{MoE}_\ell(x) = \sum_i z_{\ell,i}(x) \cdot E_{\ell,i}(x)$. 
LoRA adapts each expert's feed-forward projections via low-rank injection: $\mathbf{W}' = \mathbf{W} + \mathbf{B}\mathbf{A}$. 
Each expert contains $P = 2$ adapted projections (\texttt{up\_proj} and \texttt{down\_proj}), sharing a single expert-level active rank $r_{\ell,i}^{(t)}$ at training step $t$. 
Our goal is to construct a heterogeneous rank distribution such that the total active ranks at the end of training match LoRA: $\sum_{\ell,i} P \cdot r_{\ell,i}^{(t_{\mathrm{end}})} = L \times N \times P \times r_{\text{target}}$. 
To support dynamic growth, each expert pre-allocates physical parameter space up to $r_{\max} = 2\,r_{\text{target}}$, but only the active dimensions are used in the forward pass.

The key question is: \emph{which expert should receive the next rank dimension to achieve the largest loss reduction?} 
In MoE models, expert gradients are inherently modulated by their routing weights, and an unselected expert receives no gradient signal. 
Using a first-order approximation and a mean-field factorization (detailed in Appendix~\ref{app:motivation}), the expected benefit of expanding expert $i$'s capacity can be approximated as:
\begin{equation}
\label{eq:decomp}
\Delta \mathcal{L}_{\ell,i} \;\propto\; \underbrace{\mathbb{E}[z_{\ell,i}]}_{f_{\ell,i}:\;\text{routing frequency}} \;\cdot\; \underbrace{\mathbb{E}[q_{\ell,i} \mid z_{\ell,i} > 0]}_{g_{\ell,i}:\;\text{learning intensity when activated}}.
\end{equation}
Here, $q_{\ell,i}(x)$ denotes the effective local gradient intensity of expert $i$ when activated. 
We treat this factorization as a theoretically motivated heuristic, of which the practical value lies in capturing two distinct and complementary aspects of expert demand (Appendix~\ref{app:motivation} discusses the validity of the underlying approximation).
The two factors have clear semantics: $f_{\ell,i}$ measures the frequency with which the expert is used by the task data, and $g_{\ell,i}$ captures its average gradient activity when activated. 
Intuitively, an expert should be expanded only when it is frequently routed and still exhibits strong learning signals.

\subsection{Expert Saliency Scoring}
\label{sec:saliency}

Based on the analysis in \S\ref{sec:prelim}, we construct Expert Saliency Scoring from two complementary signals: routing frequency and rank importance.

\paragraph{Routing frequency ($f$).}
Let $B^{(t)}$ denote the set of tokens at training step $t$. We compute the step-level mean post-top-$k$ routing weight and track its exponential moving average (EMA):
\begin{equation}
\label{eq:routing}
\bar{z}^{(t)}_{\ell,i} = \frac{1}{|B^{(t)}|} \sum_{x \in B^{(t)}} z_{\ell,i}(x), \qquad
f^{(t)}_{\ell,i} = \beta \, f^{(t-1)}_{\ell,i} + (1 - \beta) \, \bar{z}^{(t)}_{\ell,i},
\end{equation}
where $\beta \in [0, 1)$ is the decay coefficient.

\paragraph{Rank importance ($g$).}
Routing frequency captures data relevance but does not guarantee that the expert is actively learning (e.g., it may have already converged). 
Therefore, we also measure learning intensity through gradient-weight products~\citep{zhang2023adaptive}. 
For the $j$-th active rank dimension, parameterized by $\mathbf{a}_j = \mathbf{A}_{\ell,i}[j,:]$ and $\mathbf{b}_j = \mathbf{B}_{\ell,i}[:,j]$, its sensitivity at step $t$ with respect to training loss $\mathcal{L}^{(t)}$ is:
\begin{equation}
\label{eq:sensitivity}
s^{(t)}_{\ell,i,j} = \left\| \frac{\partial \mathcal{L}^{(t)}}{\partial \mathbf{a}_j} \odot \mathbf{a}_j \right\|_1 \cdot \left\| \frac{\partial \mathcal{L}^{(t)}}{\partial \mathbf{b}_j} \odot \mathbf{b}_j \right\|_1.
\end{equation}
We use a multiplicative form because a rank dimension contributes meaningfully only when both its A-side and B-side carry nontrivial gradient-weight signals. 
Each dimension's sensitivity is smoothed using an EMA, then aggregated to the expert level by averaging over active dimensions. 
Let $\mathbf{m}^{(t)}_{\ell,i} \in \{0,1\}^{r_{\max}}$ be the binary mask that tracks active rank dimensions, with its $j$-th entry denoted $m^{(t)}_{\ell,i,j}$.
The expert-level rank importance is:
\begin{equation}
\label{eq:rank_imp}
g^{(t)}_{\ell,i,j} = \beta \, g^{(t-1)}_{\ell,i,j} + (1 - \beta) \, s^{(t)}_{\ell,i,j}, \qquad
g^{(t)}_{\ell,i} = \frac{1}{r^{(t)}_{\ell,i}} \sum_{j:\, m^{(t)}_{\ell,i,j}=1} g^{(t)}_{\ell,i,j}.
\end{equation}
The average of $r^{(t)}_{\ell,i}$ avoids systematically favoring experts that already have more active ranks, making $g^{(t)}_{\ell,i}$ a rank-normalized measure of learning intensity. 
Appendix~\ref{app:sensitivity_design} provides a further justification for the multiplicative form and the choice of averaging over alternative aggregation strategies.

\paragraph{Saliency score.}
We define the expert saliency score:
\begin{equation}
\label{eq:saliency}
S^{(t)}_{\ell,i} = \frac{f^{(t)}_{\ell,i} \cdot g^{(t)}_{\ell,i}}{(r^{(t)}_{\ell,i} + 1)^{\gamma}}.
\end{equation}
The numerator $f \cdot g$ serves as a practical proxy for the relative benefit of expanding expert $i$. 
The denominator assigns decreasing growth priority to experts that already hold higher ranks: an expert must exhibit stronger signals to continue receiving growth, preventing a small number of experts from monopolizing the parameter budget. 
$\gamma$ controls this effect: $\gamma = 0$ removes it entirely, causing performance degradation, while excessively large $\gamma$ suppresses meaningful differentiation (see \S\ref{sec:analysis_penalty}).

\subsection{Dynamic Rank Allocation}
\label{sec:allocation}

Using continuously updated saliency scores, \methodname~progressively grows each layer's active ranks from $r_{\text{init}}$ toward the target training budget.

\paragraph{Initialization.}
Each expert pre-allocates LoRA matrices up to $r_{\max}$ dimensions but activates only $r_{\text{init}}$ at the start. 
The binary mask $\mathbf{m}^{(t)}_{\ell,i}$ introduced in \S\ref{sec:saliency} tracks active dimensions, and the forward pass uses only the active subspace: $\mathbf{W}' = \mathbf{W} + \mathbf{B}[:,\mathbf{m}^{(t)}_{\ell,i}] \cdot \mathbf{A}[\mathbf{m}^{(t)}_{\ell,i},:]$.

\paragraph{Growth schedule.}
Rank grows every $T_{\text{grow}}$ steps within a growth window $T_{\text{window}}$. 
Let $N_{\text{grow}}$ denote the number of growth events; at each event, a fixed per-layer quota $Q = \lceil N \times (r_{\text{target}} - r_{\text{init}}) / N_{\text{grow}} \rceil$ of new ranks is distributed, reaching the target active-rank budget by the end of $T_{\text{window}}$. See Appendix~\ref{app:growth_schedule} for detailed schedule parameters.

\paragraph{Per-layer greedy allocation.}
At each growth event, \methodname~processes each layer independently. 
All $N$ experts in layer $\ell$ are sorted by $S^{(t)}_{\ell,i}$ in descending order, and the quota $Q$ is greedily distributed to the top-scoring experts. 
Per-layer allocation prevents layers with systematically larger saliency scores in practice from absorbing a disproportionate share of the budget. 
The number of new ranks assigned to each expert is capped:
\begin{equation}
\label{eq:ngrow}
n_{\text{grow}} = \min\!\Big(\lfloor (r_{\max} - r_{\text{init}}) \cdot p_{\text{grow}} \rfloor,\; r_{\max} - r^{(t)}_{\ell,i},\; Q_{\text{remain}}\Big),
\end{equation}
here $p_{\text{grow}}$ limits the maximum growth fraction per event, ensuring sufficiently gradual growth for the saliency scoring to distinguish experts with a genuinely high-demand from transient fluctuations. 
$n_{\text{grow}}$ is computed for each expert $E_{\ell,i}$ sequentially in descending saliency order, and $Q_{\text{remain}}$ denotes the remaining per-layer quota, which decreases as experts are allocated. 
After each growth event, we reset rank importance scores while preserving routing frequencies, so that newly expanded experts must justify further growth through new gradient signals rather than historical importance.

\section{Experiments}
\subsection{Experimental Setup}

\paragraph{Models.}
We evaluated \methodname~in three MoE models with different routing mechanisms and parameter sizes: \textbf{OLMoE-1B-7B} \citep{muennighoff2025olmoe}, \textbf{Qwen1.5-MoE-A2.7B} \citep{qwen_moe}, and \textbf{LLaMA-MoE-v1-3.5B (4/16)} \citep{zhu2024llama}. 

\paragraph{Training and Evaluation.}
To evaluate \methodname's ability to exploit task-specific expert specialization, we conduct independent fine-tuning on six domains, each paired with standardized benchmarks:
\textbf{Mathematical Reasoning} (\gsm~\citep{cobbe2021training}),
\textbf{Code Generation} (\humaneval~\citep{chen2021evaluating}),
\textbf{Instruction Following} (\ifeval~\citep{zhou2023instruction}),
\textbf{Medical QA} (\medmcqa~\citep{MedMCQA}),
\textbf{Machine Translation} (\wmt~\footnote{https://github.com/google-research/mt-metrics-eval}\citep{kocmi-etal-2023-findings}),
and \textbf{Legal Understanding} (\ledgar~\citep{tuggener-etal-2020-ledgar}). Training datasets and evaluation details are provided in Appendix~\ref{sec:experiment}.

\paragraph{Baselines.}
We compare \methodname~with the pretrained \textbf{Base Model} and several
strong PEFT baselines: \textbf{LoRA}~\citep{hu2022lora},
\textbf{AdaLoRA}~\citep{zhang2023adaptive}, and
\textbf{PERFT-R}~\citep{liu2024perft}.
\methodname~uses a final average active rank of
$r_{\mathrm{target}}=16$ but pre-allocates trainable parameter space up to
$r_{\max}=32$.
Accordingly, LoRA ($r=32$) serves as the primary budget-matched baseline,
with the same trainable parameter count and training-time memory footprint
as \methodname.
All methods use identical training schedules, optimization settings, and
data orders.

\begin{table*}[t]
\centering
\small
\resizebox{\linewidth}{!}{
\begin{tabular}{l c cccccc c}
\toprule
& & \textbf{Math} & \textbf{Code} & \textbf{Instruct} & \textbf{Medical} & \textbf{Translation} & \textbf{Legal} & \\
\cmidrule(lr){3-3} \cmidrule(lr){4-4} \cmidrule(lr){5-5} \cmidrule(lr){6-6} \cmidrule(lr){7-7} \cmidrule(lr){8-8}
\textbf{Method} & \textbf{\#Trainable} & \gsm & \humaneval & \ifeval & \medmcqa & \wmt & \ledgar & \textbf{Avg} \\
\midrule
\multicolumn{9}{c}{\textit{OLMoE-1B-7B}} \\
\midrule
Base             & --     & 12.9{\tiny$\pm$0.4} & 13.6{\tiny$\pm$0.3} & 15.9{\tiny$\pm$0.8} & 35.7{\tiny$\pm$0.7} & 56.6{\tiny$\pm$0.8} & 62.2{\tiny$\pm$1.2} & 32.8 \\
LoRA ($r$=32)    & 201.3M & 25.2{\tiny$\pm$0.2} & 14.8{\tiny$\pm$0.8} & 23.3{\tiny$\pm$0.4} & 40.1{\tiny$\pm$1.0} & 63.2{\tiny$\pm$1.0} & 79.4{\tiny$\pm$0.8} & 41.0 \\
AdaLoRA          & 201.3M & \underline{26.3}{\tiny$\pm$0.3} & 15.3{\tiny$\pm$0.8} & 24.1{\tiny$\pm$0.5} & \underline{41.5}{\tiny$\pm$1.0} & 63.8{\tiny$\pm$1.0} & \underline{80.1}{\tiny$\pm$0.9} & \underline{41.9} \\
PERFT-R          & 201.3M & 25.9{\tiny$\pm$0.3} & \underline{15.5}{\tiny$\pm$0.8} & \underline{24.3}{\tiny$\pm$0.5} & 41.0{\tiny$\pm$0.9} & \underline{64.0}{\tiny$\pm$1.0} & 79.6{\tiny$\pm$0.9} & 41.7 \\
\rowcolor{highlight} \textbf{\methodname~(Ours)} & 201.3M & \textbf{28.4}{\tiny$\pm$1.0} & \textbf{16.7}{\tiny$\pm$0.4} & \textbf{26.7}{\tiny$\pm$1.0} & \textbf{43.9}{\tiny$\pm$0.8} & \textbf{66.5}{\tiny$\pm$1.4} & \textbf{82.1}{\tiny$\pm$1.1} & \textbf{44.1} \\
\midrule
\multicolumn{9}{c}{\textit{Qwen1.5-MoE-A2.7B}} \\
\midrule
Base             & --     & 61.7{\tiny$\pm$1.1} & 35.0{\tiny$\pm$0.8} & 23.5{\tiny$\pm$0.6} & 43.2{\tiny$\pm$1.0} & 72.6{\tiny$\pm$1.3} & 83.2{\tiny$\pm$1.0} & 53.2 \\
LoRA ($r$=32)    & 318.5M & 64.5{\tiny$\pm$1.1} & 43.9{\tiny$\pm$0.8} & 32.0{\tiny$\pm$0.3} & 47.4{\tiny$\pm$1.2} & 78.0{\tiny$\pm$1.2} & 92.6{\tiny$\pm$0.7} & 59.7 \\
AdaLoRA          & 318.5M & 65.4{\tiny$\pm$1.0} & \underline{45.2}{\tiny$\pm$0.7} & 33.0{\tiny$\pm$0.4} & 48.0{\tiny$\pm$1.1} & \underline{79.3}{\tiny$\pm$1.1} & 93.1{\tiny$\pm$0.8} & 60.7 \\
PERFT-R          & 318.5M & \underline{65.6}{\tiny$\pm$1.0} & 44.8{\tiny$\pm$0.8} & \underline{33.3}{\tiny$\pm$0.4} & \underline{48.3}{\tiny$\pm$1.2} & 79.1{\tiny$\pm$1.1} & \underline{93.4}{\tiny$\pm$0.8} & \underline{60.8} \\
\rowcolor{highlight} \textbf{\methodname~(Ours)} & 318.5M & \textbf{67.2}{\tiny$\pm$1.1} & \textbf{46.5}{\tiny$\pm$0.6} & \textbf{35.1}{\tiny$\pm$0.2} & \textbf{50.1}{\tiny$\pm$0.8} & \textbf{80.7}{\tiny$\pm$0.6} & \textbf{94.8}{\tiny$\pm$1.0} & \textbf{62.4} \\
\midrule
\multicolumn{9}{c}{\textit{LLaMA-MoE-3.5B}} \\
\midrule
Base             & --     & 3.8{\tiny$\pm$0.8}  & 1.3{\tiny$\pm$0.5}  & 14.9{\tiny$\pm$0.5} & 25.8{\tiny$\pm$0.7} & 55.0{\tiny$\pm$1.0} & 29.6{\tiny$\pm$0.8} & 21.7 \\
LoRA ($r$=32)    & 156.8M & 12.3{\tiny$\pm$0.8} & 10.2{\tiny$\pm$0.5} & 19.7{\tiny$\pm$0.5} & 30.3{\tiny$\pm$0.5} & 63.2{\tiny$\pm$0.8} & 79.3{\tiny$\pm$1.3} & 35.8 \\
AdaLoRA          & 156.8M & \underline{13.6}{\tiny$\pm$0.7} & 10.9{\tiny$\pm$0.6} & \underline{21.2}{\tiny$\pm$0.6} & 31.1{\tiny$\pm$0.6} & \underline{64.8}{\tiny$\pm$0.9} & 79.8{\tiny$\pm$1.2} & \underline{36.9} \\
PERFT-R          & 156.8M & 13.1{\tiny$\pm$0.7} & \underline{11.1}{\tiny$\pm$0.5} & 20.5{\tiny$\pm$0.6} & \underline{31.3}{\tiny$\pm$0.5} & 64.1{\tiny$\pm$0.9} & \underline{80.0}{\tiny$\pm$1.2} & 36.7 \\
\rowcolor{highlight} \textbf{\methodname~(Ours)} & 156.8M & \textbf{15.7}{\tiny$\pm$0.5} & \textbf{13.1}{\tiny$\pm$0.6} & \textbf{23.4}{\tiny$\pm$0.8} & \textbf{32.9}{\tiny$\pm$0.9} & \textbf{65.8}{\tiny$\pm$1.1} & \textbf{81.3}{\tiny$\pm$1.2} & \textbf{38.7} \\
\bottomrule
\end{tabular}}
\caption{Task-specific adaptation results on three MoE models across six domains. 
We report mean accuracy (\%) {\tiny$\pm$std} over 3 random seeds. \textbf{Bold} indicates the best result and \underline{underline} indicates the second best.}
\label{tab:main_results}
\end{table*}

\subsection{Main Results}
\label{sec:main_results}

Table~\ref{tab:main_results} presents the main results of
\methodname{} on three pretrained MoE models.
Under matched trainable-parameter budgets, \methodname{} achieves the
best average performance on all three models, reaching
44.1, 62.4, and 38.7 on OLMoE, Qwen1.5-MoE, and LLaMA-MoE,
respectively, and outperforming the strongest baseline by
2.2, 1.6, and 1.8 points.
Moreover, \methodname{} outperforms the strongest competing baseline
in each of the 18 model--task settings, with a mean improvement of 1.8 points.
A paired $t$-test over the 18 setting-level differences confirms that
the aggregate improvement is statistically significant
($p<0.001$); detailed per-setting comparisons are provided in
Appendix~\ref{app:significance}.

The comparison with fixed-rank LoRA shows that uniformly assigning the same adaptation capacity to all experts is suboptimal for MoE fine-tuning. 
Although LoRA ($r=32$) matches \methodname{} in trainable-parameter
count, \methodname{} consistently achieves higher performance, indicating the advantage of fine-grained expert-level capacity allocation over a uniform rank configuration.
The comparison with AdaLoRA further suggests that, while dynamic rank allocation is beneficial, adjusting the budget solely based on module-level gradient importance may still be insufficient to capture the heterogeneous capacity demands of different MoE experts. 
AdaLoRA follows a top-down pruning process, starting from a higher-rank initialization and progressively pruning toward the target budget. 
Under sparse routing, low-frequency experts often receive limited and unstable gradient signals, making their importance estimates more susceptible to noise. 
In contrast, \methodname{} starts from a small active rank and expands
capacity only after expert-specific routing and learning signals have
accumulated, which enables more reliable allocation decisions.
Finally, the comparison with PERFT-R suggests that enhancing adaptation through routable adaptation modules alone may not fully address the expert-level capacity mismatch. PERFT-R focuses on the design and routing of additional adaptation
modules, whereas \methodname{} directly reallocates capacity across the
original pretrained experts.
The consistent advantage of \methodname{} therefore demonstrates the
benefit of directly performing fine-grained dynamic rank allocation over
the original experts.

\begin{table}[t]
\centering
\small
\resizebox{0.7\linewidth}{!}{
\begin{tabular}{l ccc c}
\toprule
\textbf{Variant} & \textbf{\gsm} & \textbf{\humaneval} & \textbf{\ifeval} & \textbf{Avg} \\
\midrule
w/o Routing Freq. ($g$ only) & 26.5 & 15.6 & 24.5 & 22.2 \\
w/o Rank Imp. ($f$ only)     & 27.2 & 15.3 & 24.1 & 22.2 \\
\midrule
\rowcolor{highlight} \textbf{Full \methodname} & \textbf{28.4} & \textbf{16.7} & \textbf{26.7} & \textbf{23.9} \\

\bottomrule
\end{tabular}
}
\caption{Ablation study of saliency score components on \olmoe.}
\label{tab:ablation_saliency}
\vspace{-2mm}
\end{table}

\begin{wraptable}{r}{0.6\textwidth}
\vspace{-5mm}
\centering
\small
\resizebox{0.95\linewidth}{!}{
\begin{tabular}{ll ccc c}
\toprule
\textbf{Method} & \textbf{Router} & \textbf{\gsm} & \textbf{\humaneval} & \textbf{\ifeval} & \textbf{Avg} \\
\midrule
LoRA & Frozen & 24.1 & 14.1 & 22.2 & 20.1 \\
LoRA & Unfrozen & 25.2 & 14.8 & 23.3 & 21.1 \\
\midrule
AdaLoRA & Frozen & 25.0 & 14.5 & 22.8 & 20.8 \\
AdaLoRA & Unfrozen & 26.3 & 15.3 & 24.1 & 21.9 \\
\midrule
\textbf{\methodname} & Frozen & 26.9 & 15.5 & 25.2 & 22.5 \\
\rowcolor{highlight} \textbf{\methodname} & Unfrozen & \textbf{28.4} & \textbf{16.7} & \textbf{26.7} & \textbf{23.9} \\
\bottomrule
\end{tabular}
}
\caption{Impact of router training on \olmoe.}
\label{tab:ablation_router}
\vspace{-3mm}
\end{wraptable}

\begin{figure}[t]
\centering
\begin{minipage}[t]{0.48\linewidth}
    \centering
    \includegraphics[width=\linewidth]{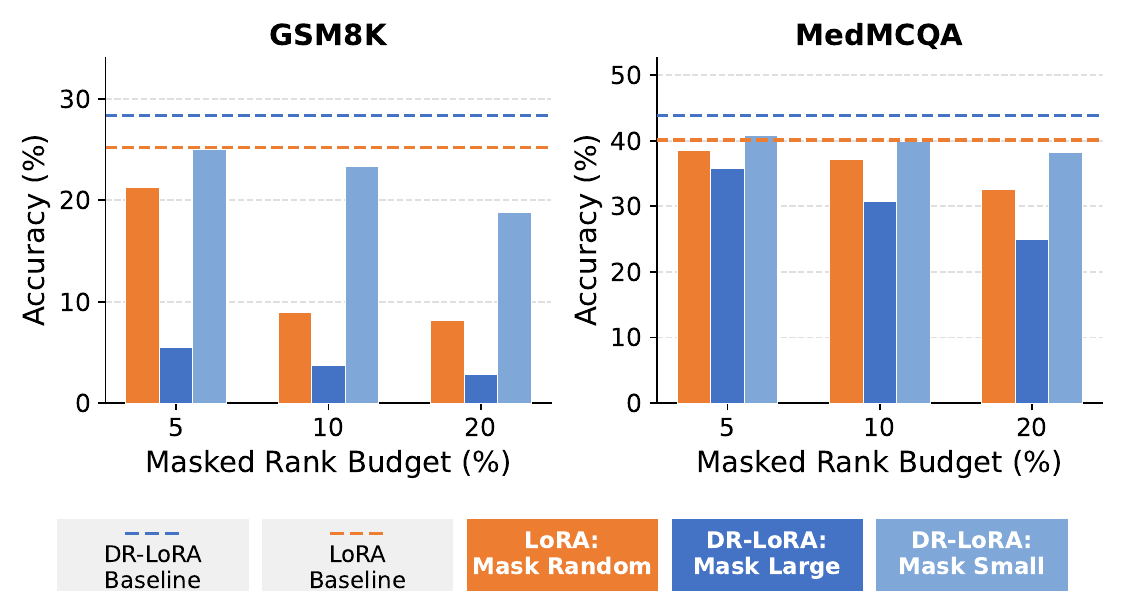}
    \caption{Expert masking on \olmoe. For \methodname, masking large-rank experts causes greater degradation than masking small-rank experts, while uniform-rank LoRA shows a less differentiated pattern.}
    \label{fig:expert_masking}
\end{minipage}
\hfill
\begin{minipage}[t]{0.48\linewidth}
    \centering
    \begin{subfigure}[b]{0.48\linewidth}
        \includegraphics[width=\linewidth]{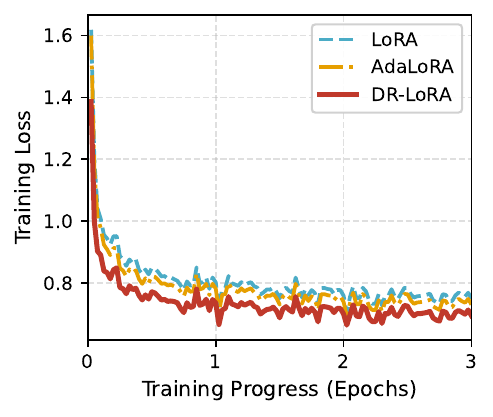}
        \caption{Training loss}
    \end{subfigure}
    \hfill
    \begin{subfigure}[b]{0.48\linewidth}
        \includegraphics[width=\linewidth]{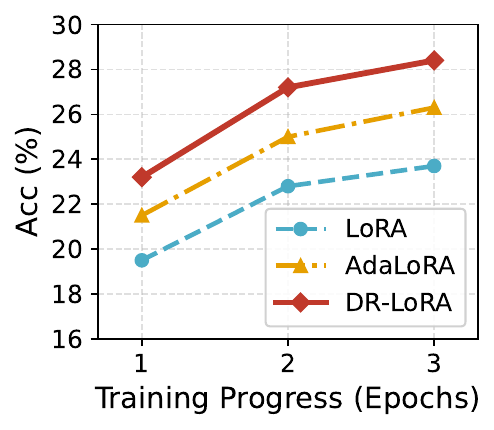}
        \caption{Accuracy}
    \end{subfigure}
    \caption{Training dynamics of \methodname~vs.\ baselines on \olmoe~fine-tuned for \gsm. \methodname~achieves consistently lower loss and higher average accuracy on \gsm~throughout training.}
    \label{fig:training_dynamics}
\end{minipage}
\vspace{-4mm}
\end{figure}

\subsection{Ablation Study} 

We conducted ablation experiments on \olmoe~to validate the key design components of \methodname. 
We report results on three representative benchmarks: \gsm, \humaneval, and \ifeval.
Table~\ref{tab:ablation_saliency} compares three variants of the saliency score: (1) full \methodname; (2) w/o routing frequency, using only gradient-based rank importance $g_{\ell,i}$; and (3) w/o rank importance, using only routing frequency $f_{\ell,i}$. 
Full \methodname~achieves the best average score of 23.9, outperforming both ablated variants by 1.7 points. 
These results indicate that routing frequency and rank importance capture complementary aspects of expert demand. 
Another question is whether the gains of \methodname~mainly result from dynamic rank allocation or simply from jointly training the MoE router. 
To better distinguish these factors, we compare LoRA, AdaLoRA, and \methodname~under both \textit{frozen} and \textit{unfrozen} router settings. 
As shown in Table~\ref{tab:ablation_router}, unfreezing the router improves all three methods, but the advantage of \methodname~cannot be attributed to router training alone. 
Notably, \methodname~with a frozen router (Avg 22.5) still outperforms LoRA with an unfrozen router (Avg 21.1). This suggests that expert-level capacity allocation provides substantial gains beyond simply updating the router under a uniform rank configuration. 
When the router is unfrozen, \methodname~achieves further gains, reaching the best overall performance.

\section{Analysis}

\subsection{Masking Analysis of the Learned Rank Allocation}

The key motivation of \methodname~is that LoRA rank allocation in MoE fine-tuning should not remain uniform across experts. 
Instead, the learned rank distribution should reflect which experts are more relevant to downstream performance on the target task. 
To examine whether the learned rank allocation exhibits such a pattern, we perform a masking analysis on \olmoe~fine-tuned for \gsm~and \medmcqa.
For each fine-tuned model, we partition experts into a large rank group (top 25\% by final LoRA rank) and a small rank group (the remaining 75\%). 
We then mask entire expert modules under three masking budgets (5\%, 10\%, and 20\% of the total rank budget per layer), and compare the resulting performance degradation. 
For the uniform-rank LoRA baseline, we randomly mask experts under the same rank budgets.
Figure~\ref{fig:expert_masking} shows a consistent pattern across both tasks: masking the large-rank expert group in \methodname~causes substantially larger degradation than masking the small-rank group. In contrast, masking experts in the uniform-rank LoRA model produce a less differentiated degradation pattern and do not exhibit the same degree of separation.
These results suggest that uniform rank allocation leads to a more even distribution of LoRA ranks across experts, whereas \methodname~learns a more differentiated rank allocation in which a smaller subset of experts contributes more strongly to downstream performance. 
This observation is consistent with our motivation that LoRA rank allocation in MoE fine-tuning should better reflect task-specific expert relevance.
 
\subsection{Analysis of Growth-Based Rank Allocation}

We analyze the growth-based design of \methodname{} from two complementary
perspectives. First, we compare growth-based and pruning-based dynamic rank
allocation. Second, we separate the benefit of progressive growth from that
of the final heterogeneous rank structure.

\paragraph{Growth versus pruning.}
Figure~\ref{fig:training_dynamics} compares the training dynamics of
fixed-rank LoRA, pruning-based AdaLoRA, and growth-based \methodname{} on
OLMoE fine-tuned for GSM8K. \methodname{} converges faster and maintains
lower training loss and higher evaluation accuracy throughout training.
Although AdaLoRA consistently outperforms fixed-rank LoRA, it remains behind
\methodname{} at all stages. AdaLoRA starts from a high-rank configuration
and progressively removes dimensions based on importance estimates. Under
sparse routing, low-frequency experts receive limited and unstable gradients,
making early pruning decisions susceptible to noise. In contrast,
\methodname{} expands capacity only after sufficient expert-specific routing
and learning signals have accumulated, making bottom-up growth better aligned
with the progressive emergence of expert importance during MoE fine-tuning.

\paragraph{Progressive growth versus fixed heterogeneous allocation.}
To separate the contribution of the final rank structure from that of progressive growth, we construct \textbf{Static-Final-Rank}. 

\begin{wraptable}{r}{0.58\textwidth}
\vspace{-5mm}
\centering
\small
\resizebox{0.98\linewidth}{!}{
\begin{tabular}{lcccc}
\toprule
\textbf{Method}
& \textbf{\gsm}
& \textbf{MedMCQA}
& \textbf{WMT23}
& \textbf{Avg.} \\
\midrule
LoRA ($r=32$)
& 25.2 & 40.1 & 63.2 & 42.8 \\
Static-Final-Rank
& 27.6 & 42.4 & 65.2 & 45.1 \\
\rowcolor{highlight}
\textbf{\methodname}
& \textbf{28.4}
& \textbf{43.9}
& \textbf{66.5}
& \textbf{46.3} \\
\bottomrule
\end{tabular}
}
\caption{
Final heterogeneous allocation versus progressive growth on \olmoe.
}
\label{tab:static_final_rank}
\vspace{-3mm}
\end{wraptable}

This variant
fixes the final per-expert rank distribution discovered by \methodname{}
from initialization and retrains the model without saliency scoring or
dynamic growth. As shown in Table~\ref{tab:static_final_rank},
Static-Final-Rank improves the average score over LoRA ($r=32$) from
42.8 to 45.1, showing that the learned heterogeneous allocation is itself
beneficial. Full \methodname{} further improves the average to 46.3,
demonstrating that progressive growth contributes an additional 1.2 points
beyond the final rank structure. This suggests that the gains arise from
both task-adaptive capacity allocation and the progressive process used to
construct it.



\begin{table}[t]
\centering
\begin{minipage}{\columnwidth}
\begin{minipage}[t]{0.47\columnwidth}
    \centering
    \vspace{0pt}
    \resizebox{\linewidth}{!}{%
    \small
    \begin{tabular}{l cccc}
        \toprule
        \textbf{Strategy} & \textbf{\gsm} & \textbf{\humaneval} & \textbf{\ifeval} & \textbf{Avg} \\
        \midrule
        LoRA & 25.2 & 14.8 & 23.3 & 21.1 \\
        \midrule
        Random       & 25.4 & 15.0 & 23.6 & 21.3 \\
        Proportional & 25.9 & 15.4 & 24.8 & 22.0 \\
        Global Greedy & 24.8 & 14.5 & 22.9 & 20.7 \\
        \rowcolor{highlight} \textbf{Per-Layer Greedy} & \textbf{28.4} & \textbf{16.7} & \textbf{26.7} & \textbf{23.9} \\
        \bottomrule
    \end{tabular}}
    \caption{Comparison of rank allocation strategies on \olmoe. All variants use the same saliency scores and parameter budget; only the allocation mechanism differs.}
    \label{tab:strategy}
\end{minipage}
\hfill
\begin{minipage}[t]{0.47\columnwidth}
    \centering
    \vspace{0pt}
    \resizebox{\linewidth}{!}{%
    \small
    \begin{tabular}{c cccc c}
        \toprule
        $\gamma$ & \textbf{\gsm} & \textbf{\humaneval} & \textbf{\ifeval} & \textbf{Avg} & \textbf{Gini} \\
        \midrule
        0 (no penalty) & 23.1 & 14.1 & 24.3 & 20.5 & 0.62 \\
        0.5            & 26.2 & 15.5 & 26.1 & 22.6 & 0.45 \\
        0.8            & 27.6 & 16.2 & 26.7 & 23.5 & 0.38 \\
        \rowcolor{highlight} \textbf{1.2 (default)} & \textbf{28.4} & \textbf{16.7} & \textbf{26.7} & \textbf{23.9} & \textbf{0.31} \\
        2.0            & 27.2 & 16.3 & 26.4 & 23.3 & 0.22 \\
        3.0            & 26.1 & 15.4 & 24.8 & 22.1 & 0.14 \\
        \midrule
        LoRA ($r$=32)  & 25.2 & 14.8 & 23.3 & 21.1 & 0.00 \\
        \bottomrule
    \end{tabular}}
    \caption{Effect of rank penalty exponent $\gamma$ on \olmoe. The Gini coefficient measures the final rank concentration across experts.}
    \label{tab:gamma_ablation}
\end{minipage}
\end{minipage}
\vspace{-5mm}
\end{table}

\subsection{Allocation Strategy Comparison}
\label{sec:strategy}
We compare five rank allocation strategies on \olmoe~under the same total training rank budget, while keeping the saliency score fixed. 
The compared strategies differ only in how the growth quota is assigned:
(I) LoRA: uses a fixed rank for all experts and does not perform dynamic allocation.
(II) Random: randomly selects experts for rank growth, without using saliency scores.
(III) Proportional: distributes the per-layer quota across all experts in proportion to their normalized saliency scores, so that higher-saliency experts receive more ranks, but lower-saliency experts may still obtain a share of the budget.
(IV) Global Greedy: pools the quota across all layers and greedily allocates it to the globally highest-scoring experts, regardless of which layer they belong to.
(V) Per-Layer Greedy (ours): allocates quota independently within each layer by greedily assigning ranks to the highest-scoring experts in that layer.
Table~\ref{tab:strategy} suggests three main observations. 
First, saliency-guided allocation is important: Random growth provides only a marginal improvement over LoRA (+0.2 avg), whereas saliency-guided strategies yield substantially larger gains, indicating that the allocation policy itself matters more than merely introducing dynamic growth. 
Second, concentrated allocation performs better than diluted allocation: Per-Layer Greedy surpasses Proportional by +1.9 avg, suggesting that spreading rank growth across many moderately salient experts is less effective than concentrating capacity on the highest-demand ones. 
Third, per-layer allocation helps avoid cross-layer imbalance: Global Greedy performs below LoRA ($-0.4$ avg), which is consistent with the tendency of some layers—often deeper ones—to accumulate systematically higher saliency scores and absorb too much of the budget. 
Per-Layer Greedy mitigates this issue by enforcing independent budgets within each layer.

\subsection{Rank Penalty Analysis}
\label{sec:analysis_penalty}

\begin{figure}[t]
\centering
\begin{minipage}[t]{0.38\columnwidth}
    \centering
    \vspace{0pt}
    \includegraphics[width=\linewidth]{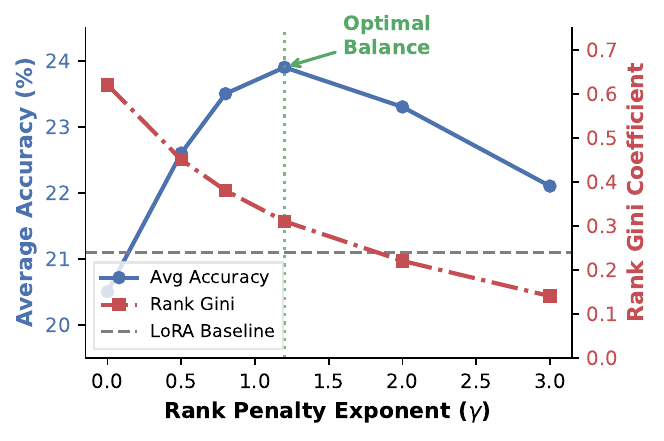}
    \captionof{figure}{Rank penalty controls the capacity concentration--performance trade-off. Left axis: average accuracy (solid); right axis: rank Gini coefficient (dashed).}
    \label{fig:gamma_gini}
\end{minipage}
\hfill
\begin{minipage}[t]{0.58\columnwidth}
    \centering
    \vspace{0pt}
    
    \begin{minipage}[t]{0.49\linewidth}
        \centering
        \vspace{0pt}
        \includegraphics[width=\linewidth]{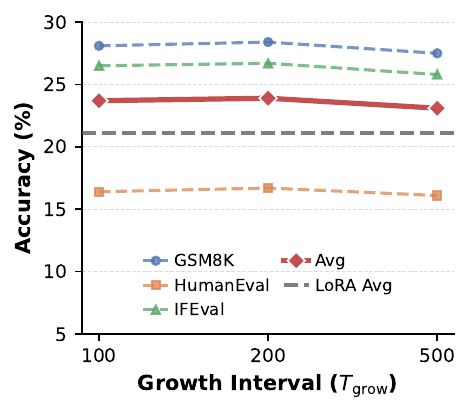}
        \centerline{\small (a) Impact of Growth Interval.}
    \end{minipage}
    \begin{minipage}[t]{0.49\linewidth}
        \centering
        \vspace{0pt}
        \includegraphics[width=0.93\linewidth]{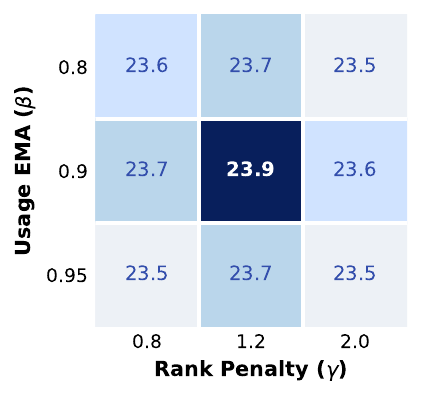}
        \centerline{\small (b) $\beta$-$\gamma$ Grid Search}
    \end{minipage}
    
    \captionof{figure}{Hyperparameter Robustness on \olmoe. \textbf{(a)} Performance across different rank growth intervals ($T_{\text{grow}}$). \textbf{(b)} A $3 \times 3$ grid search over the usage EMA coefficient ($\beta$) and rank penalty ($\gamma$).}
    \label{fig:hyperparameters}
\end{minipage}
\end{figure}

The rank penalty $(r+1)^{-\gamma}$ is introduced to discourage excessive concentration of rank growth in a small number of experts. 
To study its effect, we vary the penalty exponent $\gamma$ on \olmoe~and report both downstream performance and the Gini coefficient of the final rank distribution. 
As shown in Table~\ref{tab:gamma_ablation}, removing the penalty entirely ($\gamma{=}0$) leads to poor performance (20.5 avg), even below LoRA ($21.1$ avg), and produces the most concentrated rank distribution (Gini $=0.62$). 
As $\gamma$ increases from $0$ to $1.2$, performance improves steadily while rank concentration decreases. 
The best result is achieved at $\gamma{=}1.2$, which yields the highest average score (23.9) with a moderate degree of rank inequality (Gini $=0.31$). 
When $\gamma$ becomes too large, performance drops again, and the rank distribution becomes increasingly uniform, suggesting that overly strong regularization suppresses useful differentiation across experts.
Figure~\ref{fig:gamma_gini} further visualizes the relationship between penalty strength, rank inequality, and average task performance.
Highly concentrated allocations ($\gamma$ close to $0$) are associated with poor performance, while overly uniform allocations (large $\gamma$) also reduce performance. 
Instead, \methodname~performs best under a moderate level of rank heterogeneity, where capacity is preferentially assigned to high-demand experts without collapsing onto only a few of them. This result supports the role of the rank penalty in stabilizing expert-level capacity allocation and preventing both rank monopolization and over-regularization.

\subsection{Hyperparameter Robustness}
\label{sec:robustness}

Beyond LoRA, \methodname~introduces three additional hyperparameters: the EMA coefficient $\beta$, the rank penalty exponent $\gamma$, and the growth interval $T_{\text{grow}}$. 
We evaluate their sensitivity on \olmoe~to verify that the performance gains of \methodname~are not due to fragile hyperparameters.
As shown in Figure~\ref{fig:hyperparameters}(a), all tested growth intervals (100, 200, and 500 steps) outperform the LoRA baseline by a clear margin. 
The default setting of 200 steps achieves the highest average score (23.9), suggesting a favorable trade-off between stable saliency estimation and sufficiently responsive capacity updates.
Figure~\ref{fig:hyperparameters}(b) shows the $3 \times 3$ grid search over $\beta$ and $\gamma$. 
All configurations consistently outperform the LoRA baseline, and performance varies only within a narrow range. 
This indicates that \methodname~is not particularly sensitive to moderate changes in the saliency-scoring hyperparameters.
These results show that \methodname~is robust to its main hyperparameters. 
Therefore we use the same default settings ($\beta{=}0.9$, $\gamma{=}1.2$, $T_{\text{grow}}{=}200$) across all models and downstream tasks in the main experiments (Table~\ref{tab:main_results}), without per-task tuning.

\section{Conclusion}

In this work, we introduce \methodname, a dynamic rank allocation for LoRA in the context of PEFT for pretrained MoE LLMs.
The core idea is to move away from the standard practice of allocating a fixed, uniform rank for all expert LoRA modules during fine-tuning.
Instead, \methodname~dynamically grows expert LoRA ranks based on a combination of routing frequency and gradient-based rank importance, aiming to match adaptive capacity to each expert's downstream task relevance.
A key finding is that growth-based allocation better suits MoE fine-tuning than pruning-based alternatives, as it avoids unreliable importance estimates from sparse gradients in low-frequency experts.
The framework is evaluated on three open MoE LLMs across six benchmarks, outperforming uniform-rank baselines and state-of-the-art adaptive methods in terms of accuracy, efficiency, and robustness.
Several ablations and diagnostic visualizations support the method's claims and illuminate the dynamics of heterogeneous rank specialization.

\section*{Acknowledgements}
This work was supported in part by the Research Grants Council of the Hong Kong SAR under Grant GRF 11217823, 11216225, and Collaborative Research Fund C1042-23GF, the National Natural Science Foundation of China under Grant 62371411, InnoHK initiative, the Government of the HKSAR, Laboratory for AI-Powered Financial Technologies, and the National Key Research and Development Program of China (Grant No. 2024YFB3309702).

\clearpage
\bibliography{refs}

@inproceedings{muennighoff2025olmoe,
title={{OLM}oE: Open Mixture-of-Experts Language Models},
author={Niklas Muennighoff and Luca Soldaini and Dirk Groeneveld and Kyle Lo and Jacob Morrison and Sewon Min and Weijia Shi and Evan Pete Walsh and Oyvind Tafjord and Nathan Lambert and Yuling Gu and Shane Arora and Akshita Bhagia and Dustin Schwenk and David Wadden and Alexander Wettig and Binyuan Hui and Tim Dettmers and Douwe Kiela and Ali Farhadi and Noah A. Smith and Pang Wei Koh and Amanpreet Singh and Hannaneh Hajishirzi},
booktitle={The Thirteenth International Conference on Learning Representations},
year={2025},
url={https://openreview.net/forum?id=xXTkbTBmqq}
}

@article{jiang2024mixtral,
  title={Mixtral of experts},
  author={Jiang, Albert Q and Sablayrolles, Alexandre and Roux, Antoine and Mensch, Arthur and Savary, Blanche and Bamford, Chris and Chaplot, Devendra Singh and Casas, Diego de las and Hanna, Emma Bou and Bressand, Florian and others},
  journal={arXiv preprint arXiv:2401.04088},
  year={2024}
}

@article{dai2024deepseekmoe,
  title={Deepseekmoe: Towards ultimate expert specialization in mixture-of-experts language models},
  author={Dai, Damai and Deng, Chengqi and Zhao, Chenggang and Xu, RX and Gao, Huazuo and Chen, Deli and Li, Jiashi and Zeng, Wangding and Yu, Xingkai and Wu, Yu and others},
  journal={arXiv preprint arXiv:2401.06066},
  year={2024}
}

@article{liu2024deepseek,
  title={Deepseek-v3 technical report},
  author={Liu, Aixin and Feng, Bei and Xue, Bing and Wang, Bingxuan and Wu, Bochao and Lu, Chengda and Zhao, Chenggang and Deng, Chengqi and Zhang, Chenyu and Ruan, Chong and others},
  journal={arXiv preprint arXiv:2412.19437},
  year={2024}
}

@article{yang2025qwen3,
  title={Qwen3 technical report},
  author={Yang, An and Li, Anfeng and Yang, Baosong and Zhang, Beichen and Hui, Binyuan and Zheng, Bo and Yu, Bowen and Gao, Chang and Huang, Chengen and Lv, Chenxu and others},
  journal={arXiv preprint arXiv:2505.09388},
  year={2025}
}

@inproceedings{shazeer2017,
title={Outrageously Large Neural Networks: The Sparsely-Gated Mixture-of-Experts Layer},
author={Noam Shazeer and *Azalia Mirhoseini and *Krzysztof Maziarz and Andy Davis and Quoc Le and Geoffrey Hinton and Jeff Dean},
booktitle={International Conference on Learning Representations},
year={2017},
url={https://openreview.net/forum?id=B1ckMDqlg}
}

@article{fedus2022switch,
  title={Switch transformers: Scaling to trillion parameter models with simple and efficient sparsity},
  author={Fedus, William and Zoph, Barret and Shazeer, Noam},
  journal={Journal of Machine Learning Research},
  volume={23},
  number={120},
  pages={1--39},
  year={2022}
}

@article{team2025kimi,
  title={Kimi-vl technical report},
  author={Team, Kimi and Du, Angang and Yin, Bohong and Xing, Bowei and Qu, Bowen and Wang, Bowen and Chen, Cheng and Zhang, Chenlin and Du, Chenzhuang and Wei, Chu and others},
  journal={arXiv preprint arXiv:2504.07491},
  year={2025}
}

@inproceedings{hu2022lora,
title={Lo{RA}: Low-Rank Adaptation of Large Language Models},
author={Edward J Hu and yelong shen and Phillip Wallis and Zeyuan Allen-Zhu and Yuanzhi Li and Shean Wang and Lu Wang and Weizhu Chen},
booktitle={International Conference on Learning Representations},
year={2022},
url={https://openreview.net/forum?id=nZeVKeeFYf9}
}

@inproceedings{wang2025hmoe,
  title={Hmoe: Heterogeneous mixture of experts for language modeling},
  author={Wang, An and Sun, Xingwu and Xie, Ruobing and Li, Shuaipeng and Zhu, Jiaqi and Yang, Zhen and Zhao, Pinxue and Han, Weidong and Kang, Zhanhui and Wang, Di and others},
  booktitle={Proceedings of the 2025 Conference on Empirical Methods in Natural Language Processing},
  pages={21954--21968},
  year={2025}
}

@article{li2024mixlora,
  title={Mixlora: Enhancing large language models fine-tuning with lora-based mixture of experts},
  author={Li, Dengchun and Ma, Yingzi and Wang, Naizheng and Ye, Zhengmao and Cheng, Zhiyuan and Tang, Yinghao and Zhang, Yan and Duan, Lei and Zuo, Jie and Yang, Cal and others},
  journal={arXiv preprint arXiv:2404.15159},
  year={2024}
}

@inproceedings{dou2024loramoe,
  title={LoRAMoE: Alleviating world knowledge forgetting in large language models via MoE-style plugin},
  author={Dou, Shihan and Zhou, Enyu and Liu, Yan and Gao, Songyang and Shen, Wei and Xiong, Limao and Zhou, Yuhao and Wang, Xiao and Xi, Zhiheng and Fan, Xiaoran and others},
  booktitle={Proceedings of the 62nd Annual Meeting of the Association for Computational Linguistics (Volume 1: Long Papers)},
  pages={1932--1945},
  year={2024}
}

@inproceedings{zhang2023adaptive,
title={Adaptive Budget Allocation for Parameter-Efficient Fine-Tuning },
author={Qingru Zhang and Minshuo Chen and Alexander Bukharin and Pengcheng He and Yu Cheng and Weizhu Chen and Tuo Zhao},
booktitle={The Eleventh International Conference on Learning Representations },
year={2023},
url={https://openreview.net/forum?id=lq62uWRJjiY}
}

@inproceedings{gao2025mola,
  title={MoLA: MoE LoRA with layer-wise expert allocation},
  author={Gao, Chongyang and Chen, Kezhen and Rao, Jinmeng and Liu, Ruibo and Sun, Baochen and Zhang, Yawen and Peng, Daiyi and Guo, Xiaoyuan and Subrahmanian, VS},
  booktitle={Findings of the Association for Computational Linguistics: NAACL 2025},
  pages={5097--5112},
  year={2025}
}

@inproceedings{liu2024moe,
  title={When moe meets llms: Parameter efficient fine-tuning for multi-task medical applications},
  author={Liu, Qidong and Wu, Xian and Zhao, Xiangyu and Zhu, Yuanshao and Xu, Derong and Tian, Feng and Zheng, Yefeng},
  booktitle={Proceedings of the 47th International ACM SIGIR Conference on Research and Development in Information Retrieval},
  pages={1104--1114},
  year={2024}
}

@article{cobbe2021training,
  title={Training verifiers to solve math word problems},
  author={Cobbe, Karl and Kosaraju, Vineet and Bavarian, Mohammad and Chen, Mark and Jun, Heewoo and Kaiser, Lukasz and Plappert, Matthias and Tworek, Jerry and Hilton, Jacob and Nakano, Reiichiro and others},
  journal={arXiv preprint arXiv:2110.14168},
  year={2021}
}

@article{chen2021evaluating,
  title={Evaluating large language models trained on code},
  author={Chen, Mark},
  journal={arXiv preprint arXiv:2107.03374},
  year={2021}
}

@article{zhou2023instruction,
  title={Instruction-following evaluation for large language models},
  author={Zhou, Jeffrey and Lu, Tianjian and Mishra, Swaroop and Brahma, Siddhartha and Basu, Sujoy and Luan, Yi and Zhou, Denny and Hou, Le},
  journal={arXiv preprint arXiv:2311.07911},
  year={2023}
}

@inproceedings{liu2024dora,
  title={Dora: Weight-decomposed low-rank adaptation},
  author={Liu, Shih-Yang and Wang, Chien-Yi and Yin, Hongxu and Molchanov, Pavlo and Wang, Yu-Chiang Frank and Cheng, Kwang-Ting and Chen, Min-Hung},
  booktitle={Forty-first International Conference on Machine Learning},
  year={2024}
}

@article{hayou2024lora+,
  title={Lora+: Efficient low rank adaptation of large models},
  author={Hayou, Soufiane and Ghosh, Nikhil and Yu, Bin},
  journal={arXiv preprint arXiv:2402.12354},
  year={2024}
}

@article{liu2024perft,
  title={Perft: Parameter-efficient routed fine-tuning for mixture-of-expert model},
  author={Liu, Yilun and Ma, Yunpu and Chen, Shuo and Ding, Zifeng and He, Bailan and Han, Zhen and Tresp, Volker},
  journal={arXiv preprint arXiv:2411.08212},
  year={2024}
}

@inproceedings{wang-etal-2024-expert,
    title = "Let the Expert Stick to His Last: Expert-Specialized Fine-Tuning for Sparse Architectural Large Language Models",
    author = "Wang, Zihan  and
      Chen, Deli  and
      Dai, Damai  and
      Xu, Runxin  and
      Li, Zhuoshu  and
      Wu, Yu",
    editor = "Al-Onaizan, Yaser  and
      Bansal, Mohit  and
      Chen, Yun-Nung",
    booktitle = "Proceedings of the 2024 Conference on Empirical Methods in Natural Language Processing",
    month = nov,
    year = "2024",
    address = "Miami, Florida, USA",
    publisher = "Association for Computational Linguistics",
    url = "https://aclanthology.org/2024.emnlp-main.46/",
    doi = "10.18653/v1/2024.emnlp-main.46",
    pages = "784--801"
}

@article{pirchert2026flexmore,
  title={FlexMoRE: A Flexible Mixture of Rank-heterogeneous Experts for Efficient Federatedly-trained Large Language Models},
  author={Pirchert, Annemette Brok and Nielsen, Jacob and From, Mogens Henrik and Poech, Lukas Galke and Schneider-Kamp, Peter},
  journal={arXiv preprint arXiv:2602.08818},
  year={2026}
}

@InProceedings{MedMCQA,
  title = 	 {MedMCQA: A Large-scale Multi-Subject Multi-Choice Dataset for Medical domain Question Answering},
  author =       {Pal, Ankit and Umapathi, Logesh Kumar and Sankarasubbu, Malaikannan},
  booktitle = 	 {Proceedings of the Conference on Health, Inference, and Learning},
  pages = 	 {248--260},
  year = 	 {2022},
  editor = 	 {Flores, Gerardo and Chen, George H and Pollard, Tom and Ho, Joyce C and Naumann, Tristan},
  volume = 	 {174},
  series = 	 {Proceedings of Machine Learning Research},
  month = 	 {07--08 Apr},
  publisher =    {PMLR},
  url = 	 {https://proceedings.mlr.press/v174/pal22a.html}
}

@misc{qwen_moe,
    title = {Qwen1.5-MoE: Matching 7B Model Performance with 1/3 Activated Parameters"},
    url = {https://qwenlm.github.io/blog/qwen-moe/},
    author = {Qwen Team},
    month = {February},
    year = {2024}
}

@inproceedings{zhu2024llama,
  title={Llama-moe: Building mixture-of-experts from llama with continual pre-training},
  author={Zhu, Tong and Qu, Xiaoye and Dong, Daize and Ruan, Jiacheng and Tong, Jingqi and He, Conghui and Cheng, Yu},
  booktitle={Proceedings of the 2024 conference on empirical methods in natural language processing},
  pages={15913--15923},
  year={2024}
}

@inproceedings{tuggener-etal-2020-ledgar,
    title = "{LEDGAR}: A Large-Scale Multi-label Corpus for Text Classification of Legal Provisions in Contracts",
    author = {Tuggener, Don  and
      von D{\"a}niken, Pius  and
      Peetz, Thomas  and
      Cieliebak, Mark},
    editor = "Calzolari, Nicoletta  and
      B{\'e}chet, Fr{\'e}d{\'e}ric  and
      Blache, Philippe  and
      Choukri, Khalid  and
      Cieri, Christopher  and
      Declerck, Thierry  and
      Goggi, Sara  and
      Isahara, Hitoshi  and
      Maegaard, Bente  and
      Mariani, Joseph  and
      Mazo, H{\'e}l{\`e}ne  and
      Moreno, Asuncion  and
      Odijk, Jan  and
      Piperidis, Stelios",
    booktitle = "Proceedings of the Twelfth Language Resources and Evaluation Conference",
    month = may,
    year = "2020",
    address = "Marseille, France",
    publisher = "European Language Resources Association",
    url = "https://aclanthology.org/2020.lrec-1.155/",
    pages = "1235--1241",
    language = "eng",
    ISBN = "979-10-95546-34-4"
}

@misc{codealpaca,
  author = {Sahil Chaudhary},
  title = {Code Alpaca: An Instruction-following LLaMA model for code generation},
  year = {2023},
  publisher = {GitHub},
  journal = {GitHub repository},
  howpublished = {\url{https://github.com/sahil280114/codealpaca}},
}

@misc{yu2024metamathbootstrapmathematicalquestions,
      title={MetaMath: Bootstrap Your Own Mathematical Questions for Large Language Models}, 
      author={Longhui Yu and Weisen Jiang and Han Shi and Jincheng Yu and Zhengying Liu and Yu Zhang and James T. Kwok and Zhenguo Li and Adrian Weller and Weiyang Liu},
      year={2024},
      eprint={2309.12284},
      archivePrefix={arXiv},
      primaryClass={cs.CL},
      url={https://arxiv.org/abs/2309.12284}, 
}

@inproceedings{bojar-etal-2017-findings,
    title = "Findings of the 2017 Conference on Machine Translation ({WMT}17)",
    author = "Bojar, Ond{\v{r}}ej  and
      Chatterjee, Rajen  and
      Federmann, Christian  and
      Graham, Yvette  and
      Haddow, Barry  and
      Huang, Shujian  and
      Huck, Matthias  and
      Koehn, Philipp  and
      Liu, Qun  and
      Logacheva, Varvara  and
      Monz, Christof  and
      Negri, Matteo  and
      Post, Matt  and
      Rubino, Raphael  and
      Specia, Lucia  and
      Turchi, Marco",
    editor = "Bojar, Ond{\v{r}}ej  and
      Buck, Christian  and
      Chatterjee, Rajen  and
      Federmann, Christian  and
      Graham, Yvette  and
      Haddow, Barry  and
      Huck, Matthias  and
      Yepes, Antonio Jimeno  and
      Koehn, Philipp  and
      Kreutzer, Julia",
    booktitle = "Proceedings of the Second Conference on Machine Translation",
    month = sep,
    year = "2017",
    address = "Copenhagen, Denmark",
    publisher = "Association for Computational Linguistics",
    url = "https://aclanthology.org/W17-4717/",
    doi = "10.18653/v1/W17-4717",
    pages = "169--214"
}

@inproceedings{bojar-etal-2018-findings,
    title = "Findings of the 2018 Conference on Machine Translation ({WMT}18)",
    author = "Bojar, Ond{\v{r}}ej  and
      Federmann, Christian  and
      Fishel, Mark  and
      Graham, Yvette  and
      Haddow, Barry  and
      Huck, Matthias  and
      Koehn, Philipp  and
      Monz, Christof",
    editor = "Bojar, Ond{\v{r}}ej  and
      Chatterjee, Rajen  and
      Federmann, Christian  and
      Fishel, Mark  and
      Graham, Yvette  and
      Haddow, Barry  and
      Huck, Matthias  and
      Yepes, Antonio Jimeno  and
      Koehn, Philipp  and
      Monz, Christof  and
      Negri, Matteo  and
      N{\'e}v{\'e}ol, Aur{\'e}lie  and
      Neves, Mariana  and
      Post, Matt  and
      Specia, Lucia  and
      Turchi, Marco  and
      Verspoor, Karin",
    booktitle = "Proceedings of the Third Conference on Machine Translation: Shared Task Papers",
    month = oct,
    year = "2018",
    address = "Belgium, Brussels",
    publisher = "Association for Computational Linguistics",
    url = "https://aclanthology.org/W18-6401/",
    doi = "10.18653/v1/W18-6401",
    pages = "272--303"
}

@inproceedings{barrault-etal-2019-findings,
    title = "Findings of the 2019 Conference on Machine Translation ({WMT}19)",
    author = {Barrault, Lo{\"i}c  and
      Bojar, Ond{\v{r}}ej  and
      Costa-juss{\`a}, Marta R.  and
      Federmann, Christian  and
      Fishel, Mark  and
      Graham, Yvette  and
      Haddow, Barry  and
      Huck, Matthias  and
      Koehn, Philipp  and
      Malmasi, Shervin  and
      Monz, Christof  and
      M{\"u}ller, Mathias  and
      Pal, Santanu  and
      Post, Matt  and
      Zampieri, Marcos},
    editor = "Bojar, Ond{\v{r}}ej  and
      Chatterjee, Rajen  and
      Federmann, Christian  and
      Fishel, Mark  and
      Graham, Yvette  and
      Haddow, Barry  and
      Huck, Matthias  and
      Yepes, Antonio Jimeno  and
      Koehn, Philipp  and
      Martins, Andr{\'e}  and
      Monz, Christof  and
      Negri, Matteo  and
      N{\'e}v{\'e}ol, Aur{\'e}lie  and
      Neves, Mariana  and
      Post, Matt  and
      Turchi, Marco  and
      Verspoor, Karin",
    booktitle = "Proceedings of the Fourth Conference on Machine Translation (Volume 2: Shared Task Papers, Day 1)",
    month = aug,
    year = "2019",
    address = "Florence, Italy",
    publisher = "Association for Computational Linguistics",
    url = "https://aclanthology.org/W19-5301/",
    doi = "10.18653/v1/W19-5301",
    pages = "1--61"
}

@inproceedings{barrault-etal-2020-findings,
    title = "Findings of the 2020 Conference on Machine Translation ({WMT}20)",
    author = {Barrault, Lo{\"i}c  and
      Biesialska, Magdalena  and
      Bojar, Ond{\v{r}}ej  and
      Costa-juss{\`a}, Marta R.  and
      Federmann, Christian  and
      Graham, Yvette  and
      Grundkiewicz, Roman  and
      Haddow, Barry  and
      Huck, Matthias  and
      Joanis, Eric  and
      Kocmi, Tom  and
      Koehn, Philipp  and
      Lo, Chi-kiu  and
      Ljube{\v{s}}i{\'c}, Nikola  and
      Monz, Christof  and
      Morishita, Makoto  and
      Nagata, Masaaki  and
      Nakazawa, Toshiaki  and
      Pal, Santanu  and
      Post, Matt  and
      Zampieri, Marcos},
    editor = {Barrault, Lo{\"i}c  and
      Bojar, Ond{\v{r}}ej  and
      Bougares, Fethi  and
      Chatterjee, Rajen  and
      Costa-juss{\`a}, Marta R.  and
      Federmann, Christian  and
      Fishel, Mark  and
      Fraser, Alexander  and
      Graham, Yvette  and
      Guzman, Paco  and
      Haddow, Barry  and
      Huck, Matthias  and
      Yepes, Antonio Jimeno  and
      Koehn, Philipp  and
      Martins, Andr{\'e}  and
      Morishita, Makoto  and
      Monz, Christof  and
      Nagata, Masaaki  and
      Nakazawa, Toshiaki  and
      Negri, Matteo},
    booktitle = "Proceedings of the Fifth Conference on Machine Translation",
    month = nov,
    year = "2020",
    address = "Online",
    publisher = "Association for Computational Linguistics",
    url = "https://aclanthology.org/2020.wmt-1.1/",
    doi = "10.18653/v1/2020.wmt-1.1",
    pages = "1--55"
}

@inproceedings{kocmi-etal-2022-findings,
    title = "Findings of the 2022 Conference on Machine Translation ({WMT}22)",
    author = "Kocmi, Tom  and
      Bawden, Rachel  and
      Bojar, Ond{\v{r}}ej  and
      Dvorkovich, Anton  and
      Federmann, Christian  and
      Fishel, Mark  and
      Gowda, Thamme  and
      Graham, Yvette  and
      Grundkiewicz, Roman  and
      Haddow, Barry  and
      Knowles, Rebecca  and
      Koehn, Philipp  and
      Monz, Christof  and
      Morishita, Makoto  and
      Nagata, Masaaki  and
      Nakazawa, Toshiaki  and
      Nov{\'a}k, Michal  and
      Popel, Martin  and
      Popovi{\'c}, Maja",
    editor = {Koehn, Philipp  and
      Barrault, Lo{\"i}c  and
      Bojar, Ond{\v{r}}ej  and
      Bougares, Fethi  and
      Chatterjee, Rajen  and
      Costa-juss{\`a}, Marta R.  and
      Federmann, Christian  and
      Fishel, Mark  and
      Fraser, Alexander  and
      Freitag, Markus  and
      Graham, Yvette  and
      Grundkiewicz, Roman  and
      Guzman, Paco  and
      Haddow, Barry  and
      Huck, Matthias  and
      Jimeno Yepes, Antonio  and
      Kocmi, Tom  and
      Martins, Andr{\'e}  and
      Morishita, Makoto  and
      Monz, Christof  and
      Nagata, Masaaki  and
      Nakazawa, Toshiaki  and
      Negri, Matteo  and
      N{\'e}v{\'e}ol, Aur{\'e}lie  and
      Neves, Mariana  and
      Popel, Martin  and
      Turchi, Marco  and
      Zampieri, Marcos},
    booktitle = "Proceedings of the Seventh Conference on Machine Translation (WMT)",
    month = dec,
    year = "2022",
    address = "Abu Dhabi, United Arab Emirates (Hybrid)",
    publisher = "Association for Computational Linguistics",
    url = "https://aclanthology.org/2022.wmt-1.1/",
    doi = "10.18653/v1/2022.wmt-1.1",
    pages = "1--45"
}

@inproceedings{akhbardeh-etal-2021-findings,
    title = "Findings of the 2021 Conference on Machine Translation ({WMT}21)",
    author = "Akhbardeh, Farhad  and
      Arkhangorodsky, Arkady  and
      Biesialska, Magdalena  and
      Bojar, Ond{\v{r}}ej  and
      Chatterjee, Rajen  and
      Chaudhary, Vishrav  and
      Costa-jussa, Marta R.  and
      Espa{\~n}a-Bonet, Cristina  and
      Fan, Angela  and
      Federmann, Christian  and
      Freitag, Markus  and
      Graham, Yvette  and
      Grundkiewicz, Roman  and
      Haddow, Barry  and
      Harter, Leonie  and
      Heafield, Kenneth  and
      Homan, Christopher  and
      Huck, Matthias  and
      Amponsah-Kaakyire, Kwabena  and
      Kasai, Jungo  and
      Khashabi, Daniel  and
      Knight, Kevin  and
      Kocmi, Tom  and
      Koehn, Philipp  and
      Lourie, Nicholas  and
      Monz, Christof  and
      Morishita, Makoto  and
      Nagata, Masaaki  and
      Nagesh, Ajay  and
      Nakazawa, Toshiaki  and
      Negri, Matteo  and
      Pal, Santanu  and
      Tapo, Allahsera Auguste  and
      Turchi, Marco  and
      Vydrin, Valentin  and
      Zampieri, Marcos",
    editor = "Barrault, Loic  and
      Bojar, Ondrej  and
      Bougares, Fethi  and
      Chatterjee, Rajen  and
      Costa-jussa, Marta R.  and
      Federmann, Christian  and
      Fishel, Mark  and
      Fraser, Alexander  and
      Freitag, Markus  and
      Graham, Yvette  and
      Grundkiewicz, Roman  and
      Guzman, Paco  and
      Haddow, Barry  and
      Huck, Matthias  and
      Yepes, Antonio Jimeno  and
      Koehn, Philipp  and
      Kocmi, Tom  and
      Martins, Andre  and
      Morishita, Makoto  and
      Monz, Christof",
    booktitle = "Proceedings of the Sixth Conference on Machine Translation",
    month = nov,
    year = "2021",
    address = "Online",
    publisher = "Association for Computational Linguistics",
    url = "https://aclanthology.org/2021.wmt-1.1/",
    pages = "1--88"
}

@inproceedings{kocmi-etal-2023-findings,
    title = "Findings of the 2023 Conference on Machine Translation ({WMT}23): {LLM}s Are Here but Not Quite There Yet",
    author = "Kocmi, Tom  and
      Avramidis, Eleftherios  and
      Bawden, Rachel  and
      Bojar, Ond{\v{r}}ej  and
      Dvorkovich, Anton  and
      Federmann, Christian  and
      Fishel, Mark  and
      Freitag, Markus  and
      Gowda, Thamme  and
      Grundkiewicz, Roman  and
      Haddow, Barry  and
      Koehn, Philipp  and
      Marie, Benjamin  and
      Monz, Christof  and
      Morishita, Makoto  and
      Murray, Kenton  and
      Nagata, Masaaki  and
      Nakazawa, Toshiaki  and
      Popel, Martin  and
      Popovi{\'c}, Maja  and
      Shmatova, Mariya  and
      Suzuki, Jun",
    editor = "Koehn, Philipp  and
      Haddow, Barry  and
      Kocmi, Tom  and
      Monz, Christof",
    booktitle = "Proceedings of the Eighth Conference on Machine Translation",
    month = dec,
    year = "2023",
    address = "Singapore",
    publisher = "Association for Computational Linguistics",
    url = "https://aclanthology.org/2023.wmt-1.1/",
    doi = "10.18653/v1/2023.wmt-1.1",
    pages = "1--42"
}

@inproceedings{feng-etal-2025-comoe,
    title = "{C}o{M}o{E}: Contrastive Representation for Mixture-of-Experts in Parameter-Efficient Fine-tuning",
    author = "Feng, Jinyuan  and
      Wei, ChaoPeng  and
      Qiu, Tenghai  and
      Hu, Tianyi  and
      Pu, Zhiqiang",
    editor = "Christodoulopoulos, Christos  and
      Chakraborty, Tanmoy  and
      Rose, Carolyn  and
      Peng, Violet",
    booktitle = "Findings of the Association for Computational Linguistics: EMNLP 2025",
    month = nov,
    year = "2025",
    address = "Suzhou, China",
    publisher = "Association for Computational Linguistics",
    url = "https://aclanthology.org/2025.findings-emnlp.398/",
    doi = "10.18653/v1/2025.findings-emnlp.398",
    pages = "7533--7551",
    ISBN = "979-8-89176-335-7"
}

@misc{tang2026exploring,
title={Exploring Expert Concentration for Parameter-efficient Fine-tuning of Mixture-of-Expert {LLM}s},
author={Yiru Tang and Kun Zhou and Xin Zhao and Jing Sha and Zhichao Sheng and Shijin Wang},
year={2026},
url={https://openreview.net/forum?id=zBgjWTWgCh}
}

@inproceedings{li-etal-2026-macs,
    title = "{MACS}: Modality-Aware Capacity Scaling for Efficient Multimodal {M}o{E} Inference",
    author = "Li, Bo  and
      Wu, Chuan  and
      Zhu, Shaolin",
    editor = "Liakata, Maria  and
      Moreira, Viviane P.  and
      Zhang, Jiajun  and
      Jurgens, David",
    booktitle = "Proceedings of the 64th Annual Meeting of the {A}ssociation for {C}omputational {L}inguistics (Volume 1: Long Papers)",
    month = jul,
    year = "2026",
    address = "San Diego, California, United States",
    publisher = "Association for Computational Linguistics",
    url = "https://aclanthology.org/2026.acl-long.1012/",
    doi = "10.18653/v1/2026.acl-long.1012",
    pages = "22134--22148",
    ISBN = "979-8-89176-390-6"
}

@article{li2026mix,
  title={Mix-MoE: Improving Multilingual Machine Translation of Large Language Models through Mixed MoEs},
  author={Li, Bo and Dong, Tianyu and Zhu, Shaolin and Xiong, Deyi},
  journal={IEEE Transactions on Audio, Speech and Language Processing},
  year={2026},
  publisher={IEEE}
}
\bibliographystyle{colm2026_conference}

\clearpage
\section*{Statement on LLM Usage} 
\label{appendix:llm_usage}
All core research ideas, experimental design, data analysis, interpretation of results, and the final scientific conclusions presented in this paper were conceived and formulated by the human authors. 
The LLM's role was strictly that of an assistant for language and expression, not a contributor to the core scientific content or intellectual contributions of the work. 
The authors take full responsibility for all content in this manuscript, including any text that was assisted by an LLM.

\section*{Limitation}

Although \methodname~demonstrates consistent improvements over strong baselines, several limitations remain. 
First, the saliency score is motivated by a first-order approximation and a mean-field factorization, and should therefore be viewed as a practical allocation heuristic rather than a strict optimal estimator of expert utility. 
Its behavior under more strongly coupled routing--gradient dynamics remains to be studied further. 
Second, our experiments are conducted primarily on text MoE LLMs with top-$k$ routing. Whether the same design transfers equally well to other routing paradigms (e.g., expert-choice routing) or to multimodal MoE architectures is still an open question. 
In addition, dynamic rank growth requires pre-allocated rank space during training, which introduces extra training-time overhead. 
Understanding this trade-off more systematically at larger scales is an important direction for future work.

\appendix

\section{Motivation for the Saliency Score}
\label{app:motivation}

This appendix provides additional motivation for the saliency score design presented in \S\ref{sec:prelim} of the main paper, discusses the practical validity of the key approximation, and explains the design choices behind the rank importance measure.

\subsection{From Capacity Expansion to the \texorpdfstring{$f\!\cdot\!g$}{f*g} Factorization}
\label{app:fg}

Under a fixed total training rank budget, the core allocation decision is: which expert should receive the next rank dimension to maximize the reduction in training loss? 
We address this question using a first-order Taylor approximation.
Consider expanding the LoRA module of the expert $E_{\ell,i}$ by one rank dimension, introducing a parameter change $\Delta\boldsymbol{\theta}_{\ell,i}$. 
For an input token $x$, the first-order approximation of the loss change is:
\begin{equation}
\label{eq:app_taylor}
\Delta \mathcal{L}(x) \;\approx\; \langle \nabla_{\boldsymbol{\theta}_{\ell,i}} \mathcal{L}(x),\; \Delta\boldsymbol{\theta}_{\ell,i} \rangle.
\end{equation}
With a norm constraint on $\Delta\boldsymbol{\theta}_{\ell,i}$, the achievable first-order decrease depends on the alignment between the gradient direction and the parameter change, indicating that gradient magnitude is a natural proxy for the potential benefit of capacity expansion.

In dense models, every module processes all tokens and receives the correspondingly dense gradient signals. 
The key difference in MoE models is that expert $i$'s output is weighted by the routing weight $z_{\ell,i}(x)$ before entering subsequent computation. 
The expert's parameter gradient is inherently modulated by its routing weight:
\begin{equation}
\label{eq:app_chain}
\nabla_{\boldsymbol{\theta}_{\ell,i}} \mathcal{L}(x) \;\propto\; z_{\ell,i}(x) \cdot \mathbf{h}_{\ell,i}(x),
\end{equation}
where $\mathbf{h}_{\ell,i}(x)$ is the local gradient signal within the expert. 
When expert $i$ is not selected by the top-$k$ routing, $z_{\ell,i}(x) = 0$ and no gradient flows through it.

Treating $q_{\ell,i}(x) = \|\mathbf{h}_{\ell,i}(x)\|$ as a measure of the effective local gradient intensity when the expert is activated (consistent with the definition in \S\ref{sec:prelim}), the expected loss reduction from expanding expert $i$'s capacity can be written as:
\begin{equation}
\label{eq:app_expected}
\Delta \mathcal{L}_{\ell,i} \;\propto\; \mathbb{E}_x\!\big[z_{\ell,i}(x) \cdot q_{\ell,i}(x)\big].
\end{equation}

We decompose this expectation as follows:
\begin{equation}
\label{eq:app_decomp}
\mathbb{E}[z_{\ell,i} \cdot q_{\ell,i}] \;=\; \mathbb{E}[z_{\ell,i}] \cdot \mathbb{E}[q_{\ell,i} \mid z_{\ell,i} > 0] \;+\; \mathrm{Cov}(z_{\ell,i},\, q_{\ell,i}).
\end{equation}
Using a mean-field-style approximation that neglects the covariance term, we obtain:
\begin{equation}
\label{eq:app_meanfield}
\Delta \mathcal{L}_{\ell,i} \;\approx\; \underbrace{\mathbb{E}[z_{\ell,i}]}_{f_{\ell,i}:\;\text{routing frequency}} \;\cdot\; \underbrace{\mathbb{E}[q_{\ell,i} \mid z_{\ell,i} > 0]}_{g_{\ell,i}:\;\text{learning intensity when activated}}.
\end{equation}

\paragraph{Validity of the approximation.}
The factorization in Eq.~(\ref{eq:app_meanfield}) is exact when $z_{\ell,i}$ and $q_{\ell,i}$ are independent; in practice, it requires their correlation to be weak relative to the product of their means. 
We emphasize that we do not claim that this holds exactly, rather, the $f \cdot g$ decomposition serves as a theoretically motivated heuristic that captures two distinct and complementary aspects of expert demand. 
Its practical value lies in whether the resulting ranking of experts leads to effective allocation decisions, not in whether the covariance term is strictly zero.

However, several aspects of the training setup reduce the magnitude of the covariance term. 
\textbf{(I)} Both $f$ and $g$ are computed as exponential moving averages over many training steps (Eqs.~\ref{eq:routing}--\ref{eq:rank_imp}), which average out token-level fluctuations and substantially reduce the instantaneous correlation between routing weights and gradient intensity. 
\textbf{(II)} Rank growth decisions occur only every $T_{\text{grow}}$ steps (default: 200), so the saliency scores that drive allocation are aggregated over hundreds of mini-batches, further smoothing any residual correlation. 
\textbf{(III)} In the default training settings, the router is frozen during warmup and unfrozen afterward (\S\ref{app:train_configs}). 
When the router is trainable, the routing weights evolve and $\mathrm{Cov}(z,q)$ is, in principle, nonzero. 
However, in our training setup, the router is updated under the same optimizer schedule but represents a much smaller parameter than the expert LoRA modules. 
Therefore, the saliency statistics, which are further smoothed by EMA and updated only every $T_{\text{grow}}$ steps, evolve more gradually than the token-level routing fluctuations. 
Moreover, the ablation in Table \ref{tab:ablation_router} shows that \methodname~with a frozen router (Avg 22.5) still outperforms both LoRA (Avg 21.1) and AdaLoRA (Avg 21.9) with an unfrozen router. 
This suggests that the saliency-based allocation remains useful even in the more stable routing regime induced by a frozen router. 
Unfreezing the router provides an additional benefit on top of the dynamic rank allocation, rather than being its sole source of improvement.

\paragraph{Semantics of the multiplicative.}
\textbf{(I)} Routing frequency $f_{\ell,i}$: how frequently the expert is used in task data. Higher frequency means that increasing this expert's capacity affects more training samples.
\textbf{(II)} Conditional learning intensity $g_{\ell,i}$: how active the expert's gradients are when it is used. Even a frequently routed expert gains no benefit from additional capacity if its LoRA gradients are near zero (i.e., it has converged).
A key property of the multiplicative form is that the score naturally approaches zero when either factor is near zero. 
This prevents two problematic allocation patterns: a frequently routed but converged expert ($f$ large, $g \approx 0$) would not receive redundant capacity; a gradient-active but rarely routed expert ($f \approx 0$, $g$ large) would not receive capacity based on noisy, sparse gradient signals. 
In contrast, an additive combination $f + g$ is more likely to assign high scores in both scenarios.

\subsection{Rank Importance}
\label{app:sensitivity_design}

The per-rank sensitivity $s^{(t)}_{\ell,i,j}$ in Eq.~(\ref{eq:sensitivity}) and its expert-level aggregation $g^{(t)}_{\ell,i}$ in Eq.~(\ref{eq:rank_imp}) involve several design choices, which we justify below.

\paragraph{Multiplicative form of per-rank sensitivity.}
The sensitivity measure $s_{\ell,i,j} = \|\nabla_{\mathbf{a}_j}\mathcal{L}\odot\mathbf{a}_j\|_1 \cdot \|\nabla_{\mathbf{b}_j}\mathcal{L}\odot\mathbf{b}_j\|_1$ adapts the gradient-weight importance criterion of \citet{zhang2023adaptive} to LoRA's paired low-rank structure. 
In LoRA, the output contribution of the $j$-th rank dimension is $\mathbf{b}_j\,\mathbf{a}_j^\top \mathbf{x}$, which vanishes when either $\mathbf{a}_j$ or $\mathbf{b}_j$ is near zero. 
Accordingly, the multiplicative form ensures that a rank dimension is scored as important only when both its A-side and B-side carry nontrivial gradient-weight products. 
An additive alternative $\|\nabla_{\mathbf{a}_j}\!\odot\!\mathbf{a}_j\|_1 + \|\nabla_{\mathbf{b}_j}\!\odot\!\mathbf{b}_j\|_1$ may assign nonzero importance even when one side has effectively converged, thereby overestimating the dimension's practical contribution.

\paragraph{Why average rather than sum or max.}
The expert-level aggregation $g_{\ell,i} = \frac{1}{r^{(t)}_{\ell,i}}\sum_j g_{\ell,i,j}$ uses the mean rather than the sum or maximum over active dimensions. 
This choice is driven by the allocation context: the saliency score determines which expert should receive the \emph{next} rank dimension, so $g_{\ell,i}$ should reflect the per-dimension learning intensity (i.e., how much each existing dimension is still learning), not the total learning activity. 
Using the sum would scale linearly with $r^{(t)}_{\ell,i}$, biasing the score toward experts that already hold more ranks and creating a positive feedback loop. 
Using the maximum would make the score sensitive to outlier dimensions and vulnerable to gradient noise, particularly for low-frequency experts that accumulate only sparse gradient observations. 
The mean provides a natural rank-normalized statistic that allows fair comparison across experts with different current ranks.

\paragraph{Scale sensitivity and EMA smoothing.}
The multiplicative form of $s_{\ell,i,j}$ involves products of gradient-weight norms and is therefore sensitive to parameter scale. 
Two mechanisms mitigate this concern. 
First, the EMA smoothing in Eq.~(\ref{eq:rank_imp}) averages out step-level fluctuations, producing stable estimates of the underlying learning intensity. 
Second, $g_{\ell,i}$ enters the saliency score only through the product $f_{\ell,i} \cdot g_{\ell,i}$, and the allocation operates via \emph{relative ranking} within each layer (per-layer greedy in \S\ref{sec:allocation}), not absolute thresholds. 
Therefore, global rescaling of sensitivities has a limited effect on the allocation outcome, since the allocation depends primarily on relative ranking within each layer rather than absolute thresholds.

\subsection{Rank Penalty}
\label{app:diminishing}

The analysis in \S\ref{app:fg} considers the immediate value of adding one rank dimension, without accounting for how many ranks the expert already holds. Intuitively, the first few rank dimensions capture the dominant directions of the parameter update, while subsequent dimensions contribute progressively less, consistent with the rapid singular value decay commonly observed in low-rank matrix approximations.
To model this effect, we assume that the incremental contribution of the $r$-th rank dimension decreases with $r$. Specifically, we model the total adaptation benefit of expert $i$ as:
\begin{equation}
\label{eq:app_utility}
U_{\ell,i}(r) = \alpha_{\ell,i} \cdot \varphi(r), \qquad \alpha_{\ell,i} = f_{\ell,i} \cdot g_{\ell,i},
\end{equation}
where $\varphi(\cdot)$ is an increasing concave function ($\varphi' > 0$, $\varphi'' < 0$), indicating that each additional dimension provides a smaller incremental gain. 
In this model, the rank allocation problem becomes a discrete budget allocation problem: maximize $\sum_i \alpha_i \cdot \varphi(r_i)$ subject to $\sum_i r_i = \mathcal{B}$, where $\mathcal{B}$ is the total rank budget. 
For separable concave objectives, a greedy strategy that allocates each additional rank to the expert with the largest incremental gain is a natural and well-motivated choice. Choosing $\varphi'(r) = (r + 1)^{-\gamma}$ with $\gamma > 0$, the incremental gain for expert $i$ is:
\begin{equation}
\label{eq:app_saliency}
\frac{\partial}{\partial r_{\ell,i}}\, U_{\ell,i}(r_{\ell,i}) \;=\; \frac{f_{\ell,i} \cdot g_{\ell,i}}{(r_{\ell,i} + 1)^{\gamma}},
\end{equation}
which matches the saliency score $S_{\ell,i}$ used in Eq.~(\ref{eq:saliency}). 
The parameter $\gamma$ controls how strongly the growth priority decreases with the existing rank, which in turn determines the concentration of the final rank distribution: $\gamma = 0$ completely removes the penalty, while excessively large $\gamma$ suppresses meaningful differentiation. 
We use $\gamma = 1.2$ as the default in all experiments.

\begin{figure}[t]
\centering
\includegraphics[width=0.85\linewidth]{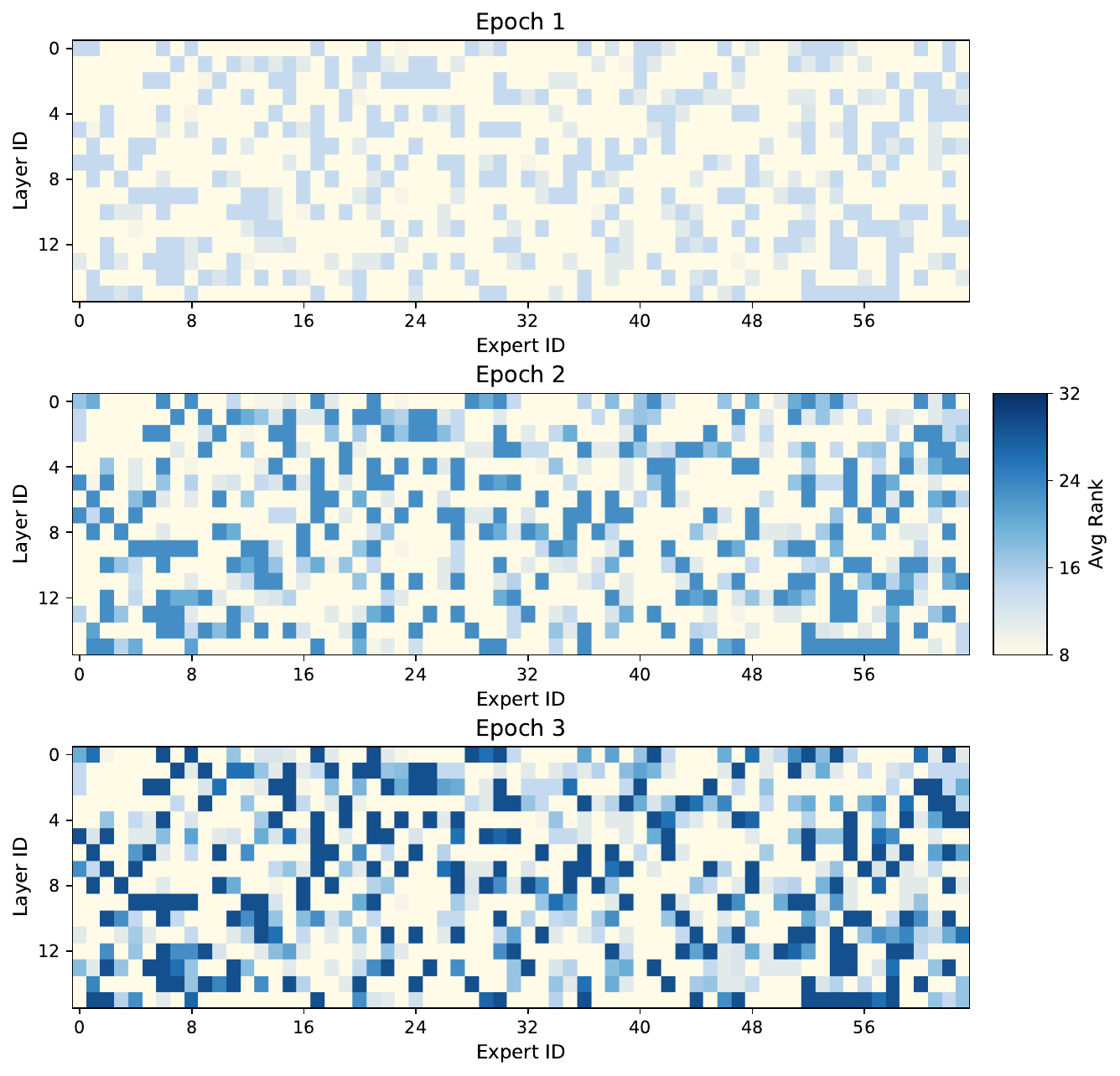}
\caption{Evolution of expert LoRA ranks during \methodname~training on \olmoe{}. Each heatmap shows the average rank per expert at different training stages, with high-rank experts (darker blue) concentrated in task-relevant positions.}
\label{fig:rank_evolution}
\end{figure}

\section{Experimental Implementation}
\label{sec:experiment}

\subsection{Training Datasets}
\label{sec:appendix_datasets}

We use six domain-specific datasets for fine-tuning:
\begin{itemize}
    \item \textbf{MetaMathQA}\footnote{https://huggingface.co/datasets/meta-math/MetaMathQA}~\citep{yu2024metamathbootstrapmathematicalquestions} is a mathematical reasoning mixture dataset. We randomly sample 60k examples.
    \item \textbf{CodeAlpaca-20k}\footnote{https://huggingface.co/datasets/sahil2801/CodeAlpaca-20k}~\citep{codealpaca} is a code instruction-following dataset containing approximately 20k samples. We use the full dataset for training.
    \item \textbf{\olmoe~SFT Mix}\footnote{https://huggingface.co/datasets/allenai/tulu-v3.1-mix-preview-4096-OLMoE}~\citep{muennighoff2025olmoe} is a general instruction-following mixture dataset used to train the \olmoe~model. We randomly sample 60k examples.
    \item \textbf{MedMCQA}\footnote{https://github.com/MedMCQA/MedMCQA}~\citep{MedMCQA} is a large-scale medical multiple-choice QA dataset. We randomly sample 60k examples from the train split.
    \item \textbf{WMT-DA-Human-Evaluation}\footnote{https://huggingface.co/datasets/RicardoRei/wmt-da-human-evaluation} is a machine translation dataset containing DA human annotations from WMT News Translation shared tasks (2017--2022)~\citep{bojar-etal-2017-findings,bojar-etal-2018-findings,barrault-etal-2019-findings,barrault-etal-2020-findings,akhbardeh-etal-2021-findings,kocmi-etal-2022-findings}. We sample 60k examples across three language pairs (en-cs, en-de, en-zh, 20k each).
    \item \textbf{LEDGAR}\footnote{https://huggingface.co/datasets/coastalchp/ledgar}~\citep{tuggener-etal-2020-ledgar} is a multilabel corpus of legal provisions in contracts. We reformat it as a multiple-choice QA task and train on the full 60k examples.
\end{itemize}

\subsection{Training Configurations}
\label{app:train_configs}

\subsubsection{Training Hyperparameters}

Table~\ref{tab:hyperparameters} presents the complete hyperparameter settings for all experiments.

\begin{table}[t]
\centering
\small
\begin{tabular}{lc}
\toprule
\textbf{Hyperparameter} & \textbf{Value} \\
\midrule
Initial rank ($r_{\text{init}}$) & 8 \\
Final Avg rank ($r_{\text{target}}$) & 16 \\
Maximum rank ($r_{\text{max}}$) & 32 \\
Growth interval ($T_{\text{grow}}$) & 200 \\
Module growth fraction ($p_{\text{grow}}$) & 0.1 \\
Usage EMA coefficient ($\beta$) & 0.9 \\
Rank penalty exponent ($\gamma$) & 1.2 \\
Learning rate & $2 \times 10^{-5}$ \\
LR scheduler & Linear \\
Warmup ratio & 0.03 \\
Weight decay & 0.0 \\
Optimizer & AdamW (fused) \\
LoRA $\alpha$ & $2 \times r$ \\
Epochs & 3 \\
Total steps & 3,750 \\
Micro-batch size & 3 \\
Gradient accumulation & 4 \\
Effective batch size & 48 \\
Max sequence length & 512 - 4096 (depending on the training datasets) \\
\bottomrule
\end{tabular}
\caption{Complete hyperparameter settings for all experiments.}
\label{tab:hyperparameters}
\end{table}

\subsubsection{Growth Schedule and Router Training}
\label{app:growth_schedule}

\noindent\textbf{Growth window ($T_{\text{window}}$):} Rank growth begins after the learning rate warmup phase ($t_{\text{warmup}}$) and stops 200 steps before training completion ($t_{\text{end}}$), and a growth window can therefore be calculated as $T_{\text{window}}=t_{\text{end}}-t_{\text{warmup}}$. This ensures that ranks activated at the final growth step still receive at least 200 steps of training before the process concludes, allowing newly added capacity to be sufficiently optimized.

\noindent\textbf{Layer synchronization:} All layers grow simultaneously at each growth event. The per-layer quota is computed as $Q = \lceil N \times (r_{\text{target}} - r_{\text{init}}) / N_{\text{grow}} \rceil$. Taking \olmoe~as an example, $N{=}128$ is the number of LoRA modules per layer (64 experts $\times$ 2 projections) and $N_{\text{grow}}$ is the number of scheduled growth events, determined by $T_{\text{window}}$ and $T_{\text{grow}}$ as $N_{\text{grow}} = \lfloor T_{\text{window}} / T_{\text{grow}} \rfloor$.

\noindent\textbf{Router training schedule:} The MoE router remains frozen during the warmup phase to stabilize LoRA training. After warmup, the router is unfrozen and trained jointly with LoRA modules until training completion. This allows the router to adapt to the evolved expert capabilities from dynamic rank allocation.

\subsubsection{Computational Infrastructure}
All experiments are conducted on a single server with 4$\times$ NVIDIA L40S GPUs (48GB each). We use DeepSpeed ZeRO-2 for distributed training with bfloat16 mixed precision, and Flash Attention 2 for memory-efficient attention computation.

\begin{table}[t]
\centering
\small
\begin{tabular}{llcccc}
\toprule
\textbf{Domain} & \textbf{Benchmark} & \textbf{Metric} & \textbf{Few-shot} & \textbf{Split} & \textbf{Size} \\
\midrule
Math            & \gsm              & Accuracy   & 8-shot & test & 1,319 \\
Code            & \humaneval          & Pass@1     & 0-shot & test & 164 \\
IF              & \ifeval             & Accuracy   & 0-shot & test & 541 \\
Med QA          & \medmcqa            & Accuracy   & 0-shot & test  & 6,150 \\
Machine Translation & \wmt          & COMET      & 0-shot & test & 4,697 \\
Legal Understanding & \ledgar         & Accuracy   & 0-shot & test & 10,000 \\
\bottomrule
\end{tabular}
\caption{Evaluation settings for all benchmarks.}
\label{tab:eval_settings}
\end{table}

\subsection{Evaluation Settings}
\label{app:eval_protocol}
All evaluations are conducted using vLLM for efficient inference, except for \llamamoe, which is not supported by vLLM and is therefore evaluated using the HuggingFace Transformers backend. All evaluations are conducted with 3 independent runs using random seeds \{42, 123, 456\} and sampling temperature $\tau = 0.2$. Table~\ref{tab:eval_settings} summarizes the evaluation configuration for each benchmark.

\section{Additional Experimental Results}

\subsection{Statistical Significance of the Main Results}
\label{app:significance}

To further assess the consistency of the improvements reported in
Table~\ref{tab:main_results}, we conduct statistical tests at both the
individual-setting and aggregate levels.
For each of the 18 model--task settings, we compare \methodname{} with
the best-performing competing baseline in that setting using a paired
$t$-test over three matched random seeds.
Table~\ref{tab:significance} reports the corresponding mean scores,
performance differences, and $p$-values.
We additionally perform an aggregate paired $t$-test by treating the
\methodname{} score and its strongest-baseline counterpart in each
model--task setting as one paired observation.

\begin{table}[t]
    \centering
    \small
    \setlength{\tabcolsep}{4.5pt}
    \begin{tabular}{lllccc}
        \toprule
        Model & Task & Best Baseline & \methodname{} & Diff. & $p$-value \\
        \midrule
        OLMoE
        & GSM8K
        & AdaLoRA (26.3)
        & 28.4
        & +2.1
        & $0.039^{*}$ \\

        & HumanEval
        & PERFT-R (15.5)
        & 16.7
        & +1.2
        & 0.068 \\

        & IFEval
        & PERFT-R (24.3)
        & 26.7
        & +2.4
        & $0.022^{*}$ \\

        & MedMCQA
        & AdaLoRA (41.5)
        & 43.9
        & +2.4
        & $0.035^{*}$ \\

        & WMT23
        & PERFT-R (64.0)
        & 66.5
        & +2.5
        & 0.213 \\

        & LEDGAR
        & AdaLoRA (80.1)
        & 82.1
        & +2.0
        & $0.009^{**}$ \\
        \midrule

        Qwen1.5-MoE
        & GSM8K
        & PERFT-R (65.6)
        & 67.2
        & +1.6
        & $0.028^{*}$ \\

        & HumanEval
        & AdaLoRA (45.2)
        & 46.5
        & +1.3
        & $0.041^{*}$ \\

        & IFEval
        & PERFT-R (33.3)
        & 35.1
        & +1.8
        & $0.007^{**}$ \\

        & MedMCQA
        & PERFT-R (48.3)
        & 50.1
        & +1.8
        & 0.259 \\

        & WMT23
        & AdaLoRA (79.3)
        & 80.7
        & +1.4
        & 0.069 \\

        & LEDGAR
        & PERFT-R (93.4)
        & 94.8
        & +1.4
        & $0.016^{*}$ \\
        \midrule

        LLaMA-MoE
        & GSM8K
        & AdaLoRA (13.6)
        & 15.7
        & +2.1
        & $0.008^{**}$ \\

        & HumanEval
        & PERFT-R (11.1)
        & 13.1
        & +2.0
        & $0.013^{*}$ \\

        & IFEval
        & AdaLoRA (21.2)
        & 23.4
        & +2.2
        & $0.012^{*}$ \\

        & MedMCQA
        & PERFT-R (31.3)
        & 32.9
        & +1.6
        & 0.059 \\

        & WMT23
        & AdaLoRA (64.8)
        & 65.8
        & +1.0
        & 0.098 \\

        & LEDGAR
        & PERFT-R (80.0)
        & 81.3
        & +1.3
        & $0.019^{*}$ \\
        \midrule

        \multicolumn{3}{l}{Aggregate across 18 settings}
        & --
        & \textbf{+1.8}
        & $\mathbf{<0.001}$ \\
        \bottomrule
    \end{tabular}
    \caption{
        Statistical comparison between \methodname{} and the
        best-performing competing baseline in each model--task setting.
        Reported method scores are averages over three random seeds.
        Per-setting $p$-values are obtained using paired $t$-tests over
        the three matched seeds.
        The aggregate test treats the 18 model--task settings as paired
        observations.
        ${}^{*}p<0.05$ and ${}^{**}p<0.01$.
    }
    \label{tab:significance}
\end{table}

As shown in Table~\ref{tab:significance}, \methodname{} outperforms the
best-performing competing baseline in all 18 model--task settings, with
performance gains ranging from 1.0 to 2.5 points.
The mean improvement across settings is 1.8 points, and the aggregate
paired test confirms that the improvement is statistically significant
($p<0.001$).
At the individual-setting level, 12 of the 18 comparisons are significant
at $p<0.05$.
The remaining comparisons are primarily associated with WMT23 and
MedMCQA, where the differences among the strongest methods are relatively
small.
Nevertheless, all setting-level differences remain positive, indicating
that the aggregate improvement is not driven by a small subset of models
or downstream tasks.

\subsection{Initial Rank Ratio Analysis}
\label{sec:appendix_rinit}

\begin{table}[t]
\centering
\small

\resizebox{0.75\columnwidth}{!}{

\begin{tabular}{l c ccc c}
\toprule
$r_{\text{init}}$ & Growth Budget & \textbf{\gsm} & \textbf{\humaneval} & \textbf{\ifeval} & \textbf{Avg} \\
\midrule
4 ($r_{\text{target}}/4$) & 12 per expert & 26.8 & 15.6 & 24.9 & 22.4 \\
\rowcolor{highlight} \textbf{8 ($r_{\text{target}}/2$)} & \textbf{8 per expert} & \textbf{28.4} & \textbf{16.7} & \textbf{26.7} & \textbf{23.9} \\
12 ($3r_{\text{target}}/4$) & 4 per expert & 27.3 & 15.9 & 25.6 & 22.9 \\
\midrule
LoRA ($r{=}32$, fixed) & 0 & 25.2 & 14.8 & 23.3 & 21.1 \\
\bottomrule
\end{tabular}
}
\caption{Impact of initial rank $r_{\text{init}}$ on \olmoe. All settings outperform LoRA (21.1 avg). $r_{\text{init}}{=}r_{\text{target}}/2$ provides the best trade-off.}
\label{tab:rinit}
\end{table}

The initial active rank $r_{\text{init}}$ determines each expert's starting capacity before dynamic growth begins. 
A smaller $r_{\text{init}}$ allows more room for saliency-guided differentiation but risks under-training early; a larger $r_{\text{init}}$ ensures stable early training but limits the growth budget for reallocation. 
We evaluate three settings on \olmoe~with fixed $r_{\text{target}}{=}16$ and $r_{\text{max}}{=}32$. 
All three configurations outperform LoRA, confirming robustness to this setting. 
The default $r_{\text{init}}{=}r_{\text{target}}/2$ delivers the best performance: it reserves half the budget for dynamic reallocation while providing enough initial capacity for stable early-stage learning. 
Too small ($r_{\text{target}}/4$): experts start with minimal capacity, causing slower convergence early before growth events begin. 
Too large ($3r_{\text{target}}/4$): the limited growth budget (only 4 ranks per expert) restricts achievable heterogeneity, reducing the benefit of dynamic allocation.

\subsection{Expert Rank Evolution Visualization}

To illustrate how \methodname~dynamically constructs heterogeneous rank distributions, we visualize the evolution of expert LoRA ranks throughout training in Figure~\ref{fig:rank_evolution}. At the early stage (Epoch 1), most experts remain at the initial rank $r_{\text{init}}=8$ (shown in light cream), with only a few high-saliency experts receiving additional capacity (darker colors). By mid-training (Epoch 2), a clear heterogeneous pattern emerges as \methodname~progressively allocates ranks to task-relevant experts based on routing frequency and rank importance. At the final stage (Epoch 3), the rank distribution becomes highly differentiated, with some experts reaching the maximum rank $r_{\text{max}}=32$ (deep navy blue) while others remain at lower ranks, reflecting their varying importance to the target task. This progression demonstrates \methodname~'s ability to automatically discover and amplify task-relevant experts through dynamic capacity allocation, forming a task-adaptive structure without manual intervention.

\section{Computational Cost Analysis}
\label{app:compute_analysis}

\subsection{Memory Analysis}
\label{app:memory_analysis}
We empirically measure the GPU memory footprint of \methodname~under our experimental 
configuration (\olmoe, 4$\times$L40S GPUs, DeepSpeed ZeRO-2, bfloat16 mixed precision). 
As shown in Table~\ref{tab:memory}, \methodname~ incurs \textbf{identical GPU memory 
usage to LoRA ($r{=}32$)}, and introduces approximately 2 GB overhead per GPU 
compared to LoRA ($r{=}16$).

\begin{table}[t]
\centering
\small
\begin{tabular}{lcc}
\toprule
\textbf{Method} & \textbf{Memory / GPU} & \textbf{Overhead vs.\ LoRA ($r{=}16$)} \\
\midrule
LoRA ($r{=}16$)  & $\sim$38 GB & --- \\
LoRA ($r{=}32$)  & $\sim$40 GB & $+$2 GB \\
\rowcolor{highlight} \methodname~($r_{\text{init}}{=}8, r_{\text{target}}{=}16$) & $\sim$40 GB & $+$2 GB \\
\bottomrule
\end{tabular}
\caption{Per-GPU memory usage under ZeRO-2 training on \olmoe~with 4$\times$L40S GPUs. 
\methodname~matches LoRA ($r{=}32$) exactly, as both allocate identical parameter space.}
\label{tab:memory}
\end{table}

This equivalence is expected: \methodname~pre-allocates LoRA parameter space up to 
$r_{\text{max}}{=}32$ for all experts, which is identical to the parameter count of 
LoRA ($r{=}32$). Under ZeRO-2, parameters and gradients (bf16) are replicated across 
devices while optimizer states (fp32) are sharded. The 2 GB overhead relative to LoRA 
($r{=}16$) thus reflects the additional 100.6M parameters allocated for the reserved 
rank dimensions ($r_{\text{max}} - r_{\text{target}} = 16$ dimensions per expert). Crucially, this memory footprint is identical to the primary baseline we compare 
against (LoRA $r{=}32$), meaning \methodname~achieves its performance gains without 
requiring any additional memory resources beyond LoRA at the matched parameter 
budget.

\subsection{Training Time Analysis}
\label{sec:training_efficiency}
We measure wall-clock training time using 4$\times$L40S GPUs across two training scenarios:
mathematical reasoning, where we train \olmoe~for three complete epochs on the MetaMathQA dataset 
and evaluate on \gsm, and code generation, where we train on CodeAlpaca-20K and evaluate 
on HumanEval. Results are summarized in Table~\ref{tab:training_time}.
\begin{table}[t]
\centering
\small
\begin{tabularx}{\columnwidth}{llXXXX}
\toprule
\textbf{Task} & \textbf{Method} & \textbf{Time} & \textbf{Overhead} & \textbf{Acc} & \textbf{$\Delta\text{Acc}$} \\
\midrule
\multirow{3}{*}{\gsm}
 & Base                                                        & -      & -    & 12.9 & -      \\
 & LoRA ($r=32$)                                               & 9.6h   & -    & 25.2 & +12.3  \\
 \rowcolor{highlight} & \textbf{\methodname~($r_{init}=8, r_{target}=16$)}          & \textbf{10.1h} & \textbf{5\%} & \textbf{28.4} & \textbf{+15.5} \\
\midrule
\multirow{3}{*}{HumanEval}
 & Base                                                        & -      & -    & 13.6 & -      \\
 & LoRA ($r=32$)                                               & 5.5h   & -    & 14.8 & +1.2   \\
 \rowcolor{highlight} & \textbf{\methodname~($r_{init}=8, r_{target}=16$)}          & \textbf{5.7h}  & \textbf{4\%} & \textbf{16.7} & \textbf{+3.1} \\
\bottomrule
\end{tabularx}
\caption{Wall-clock training time comparison across two tasks. Overhead is relative to LoRA 
($r=32$). $\Delta\text{Acc}$ denotes accuracy gain over the base model.}
\label{tab:training_time}
\end{table}
\methodname~incurs a modest training time overhead of 5\% and 4\% over LoRA ($r=32$) on 
MetaMathQA and CodeAlpaca-20K respectively. Yet despite this near-equivalent computational 
cost, \methodname~achieves substantially larger performance gains: +15.5 points on \gsm~
versus +12.3 for LoRA ($r=32$), and +3.1 points on HumanEval versus +1.2 respectively. 
This consistent pattern across both mathematical reasoning and code generation demonstrates 
that dynamic rank allocation yields substantially greater performance gains than simply 
scaling up a static rank. The overhead is attributable to dynamic rank allocation mechanisms, including importance scoring, expert usage tracking, and periodic rank growth.

\subsection{FLOPs Analysis}
We analyze the computational cost of \methodname~in terms of floating-point operations 
(FLOPs) during training. The total FLOPs consist of base expert computation and LoRA 
adaptation. For a single forward pass, base expert FLOPs are:
\begin{equation}
\text{FLOPs}_{\text{base}} = 4BLK \cdot d_m \cdot d_e
\end{equation}
where $B{=}4096$ is the effective batch size, $L{=}16$ is the number of layers, $K{=}8$ 
is the number of activated experts per layer, $d_m{=}2048$ is the hidden dimension, and 
$d_e{=}1024$ is the expert dimension. LoRA adds:
\begin{equation}
\text{FLOPs}_{\text{LoRA}} = 8BLK \cdot d_e \cdot r
\end{equation}
where $r$ is the LoRA rank and each expert has two LoRA modules (\texttt{up\_proj} and 
\texttt{down\_proj}).
Table~\ref{tab:flops_comparison} presents the FLOPs 
analysis for different methods. \methodname~has identical LoRA FLOPs to LoRA ($r{=}32$), 
as we use the maximum rank for a conservative upper-bound estimate. Since the base expert 
computation dominates total FLOPs (98.5\%), the overall increase over the no-LoRA baseline 
is only 3.1\%.
\begin{table}[t]
\centering
\small
\begin{tabular}{lcccc}
\toprule
\textbf{Method} & \textbf{Base} & \textbf{LoRA} & \textbf{Total} & \textbf{Rel.} \\
\midrule
No LoRA          & 4398.0 & ---   & 4398.0 & 1.000$\times$ \\
LoRA ($r{=}32$)  & 4398.0 & 137.4 & 4535.5 & 1.031$\times$ \\
\rowcolor{highlight} \methodname~($r_{\text{init}}{=}8, r_{\text{target}}{=}16$) & 4398.0 & 137.4 & 4535.5 & 1.031$\times$ \\
\bottomrule
\end{tabular}
\caption{FLOPs per sample (GFLOPs) for forward pass. Base expert computation dominates 
(97--99\%), making LoRA's contribution minimal. \methodname's LoRA FLOPs are 
computed using the maximum rank ($r{=}32$) as a conservative upper bound.}
\label{tab:flops_comparison}
\end{table}

\clearpage

\end{document}